\pdfoutput=1

\documentclass[11pt,a4paper]{article}

\usepackage{authblk}

\usepackage[acceptedWithA]{tacl2021v1}

\usepackage{times}
\usepackage{latexsym}

\usepackage{enumitem}
\usepackage[scientific-notation=true]{siunitx}
\usepackage{latexsym}
\usepackage{times}
\usepackage{float}
\usepackage{footnote}
\usepackage{caption, subcaption}
\usepackage{graphicx}
\usepackage{mathtools}
\usepackage{amsmath}
\usepackage{amssymb}
 \usepackage{booktabs}
 \usepackage{makecell}
 \usepackage{multirow}
 \usepackage{tabularx,stackengine}
\makesavenoteenv{tabular}
\makesavenoteenv{table}
\usepackage{graphicx}
\usepackage{hhline}
\usepackage{calc}
\usepackage{colortbl}
\usepackage{latexsym}
\usepackage{standalone}
\usepackage{url}
\usepackage{tikz}
\usepackage{soul}
\usepackage{fixltx2e}
\usepackage{color}
\usetikzlibrary{arrows,arrows.meta,decorations.pathmorphing,backgrounds,positioning,fit,petri,calc,math}
\usepackage{pgfmath}
\usepackage{courier}
\usepackage{comment}

\usepackage{tcolorbox}
\usepackage{adjustbox}

\newtcbox{\mybox}[1][red]{on line,
arc=4pt,colback=#1!10!white,colframe=white,
before upper=\strut,boxrule=0pt,
boxsep=0pt,left=2pt,right=4pt,top=-1pt,bottom=-1pt}

\newcommand{\upbox}[1]{\raisebox{1pt}{\mybox[green]{{\scriptsize{$\uparrow${#1}}}}}}

\DeclareMathOperator{\lon}{lon}
\DeclareMathOperator{\lat}{lat}

\DeclareMathOperator{\mt}{mt}
\DeclareMathOperator{\mlm}{mlm}
\DeclareMathOperator{\geo}{geo}

\newcommand{\ftg}{FT-Geoloc}
\newcommand{\zsg}{ZS-Geoloc}
\newcommand{\ftl}{FT-Lang}
\newcommand{\zsl}{ZS-Lang}
\newcommand{\zsd}{ZS-Dialect}

\newcommand{\squeezeup}{\vspace{-2.5mm}}

\usepackage[T1]{fontenc}

\usepackage[utf8]{inputenc}

\usepackage{microtype}

\title{Geographic Adaptation of Pretrained Language Models}

\makeatletter
\renewcommand\AB@affilsepx{\hspace{1em} \protect\Affilfont}
\makeatother

\renewcommand*{\Affilfont}{\normalsize\normalfont}

\newcommand{\gadas}{\mbox{GeoAda-S}}
\newcommand{\gadaw}{\mbox{GeoAda-W}}
\newcommand{\vada}{\mbox{MLMAda}}

\DeclareRobustCommand{\first}[1]{{\textbf{#1}}}
\DeclareRobustCommand{\second}[1]{{\underline{#1}}}

\author[1-3]{Valentin Hofmann}
\author[4]{Goran Glava\v{s}}
\author[5,6]{Nikola Ljube\v{s}i\'c}
\author[2]{\\Janet B. Pierrehumbert}
\author[3]{Hinrich Sch\"utze}

\affil[1]{Allen Institute for AI}
\affil[2]{University of Oxford}
\affil[3]{LMU Munich\protect\\\hspace{-1.25em}}
\affil[4]{CAIDAS, University of Würzburg}
\affil[5]{Jo\v{z}ef Stefan Institute}
\affil[6]{University of Ljubljana\protect\\\texttt{valentinh@allenai.org}}

\begin{document}
\maketitle
\begin{abstract}
While pretrained language models (PLMs) have been shown to possess a plethora of linguistic knowledge, the existing body of research has largely neglected extralinguistic knowledge, which is generally difficult to obtain by pretraining on text alone.
Here, we contribute to closing this gap by examining \emph{geolinguistic} knowledge, i.e., knowledge about geographic variation in language.
We introduce \emph{geoadaptation}, an intermediate training step that couples language
modeling with geolocation prediction in a multi-task
learning setup. We geoadapt four PLMs, covering language groups from three geographic areas, and evaluate them on five different tasks: fine-tuned (i.e., supervised)  geolocation
prediction, zero-shot (i.e., unsupervised) geolocation
prediction, fine-tuned language identification, zero-shot language identification, and zero-shot prediction of dialect
features. Geoadaptation is very
successful at injecting geolinguistic knowledge into the PLMs: 
the geoadapted PLMs consistently outperform PLMs adapted using only
language modeling (by especially wide margins on zero-shot prediction tasks), and we obtain new state-of-the-art results on two benchmarks for geolocation prediction and language identification. Furthermore, we show that the effectiveness of geoadaptation stems from its ability to 
\emph{geographically retrofit} the representation space of the PLMs.
\end{abstract}

\section{Introduction}

The default tool for the majority of NLP tasks is now \textit{de facto} pretrained language models (PLMs; \citealp[\textit{inter alia}]{Devlin.2019,Liu.2019b,Radford.2019,Brown.2020,Clark.2020,Raffel.2020,Chowdhery2022,Hoffmann2022,Touvron.2023}), which are trained using language modeling objectives on large text corpora. Despite the conceptual simplicity of language modeling, pretraining induces complex forms of linguistic knowledge in PLMs, at various levels \cite{rogers2020primer, Mahowald2023}: 
morphological \citep{Edmiston.2020, Hofmann.2020c, Weissweiler.2023}, lexical \cite{Ethayarajh.2019,vulic2020probing}, syntactic \citep{Hewitt.2019, Jawahar.2019, Wei.2021, Weissweiler.2022},
and semantic \citep{Wiedemann.2019, Ettinger.2020}. This general linguistic knowledge is then (re-)shaped for concrete tasks via fine-tuning, i.e., supervised training on task-specific labeled data.

Humans, however, additionally make use of a rich spectrum of \emph{extralinguistic} features when they learn and process language, 
including gender \citep{Lass.1979}, ethnicity \citep{Trent.1995}, and geography \citep{Clopper.2004}. Despite the growing awareness for the importance of such factors in NLP \citep{Hovy.2021}, extralinguistic features have been typically introduced in the fine-tuning phase so far, i.e., when specializing PLMs for a concrete task \cite[e.g.,][]{Rosin.2022b}. This prevents PLMs from forming generalizable representations the way humans do, impeding the exploitation of extralinguistic knowledge for tasks other
than the fine-tuning task itself.

In this work, we focus on geographic knowledge, and more specifically \textit{geolinguistic} knowledge, i.e., knowledge about geographic variation in language -- the most salient type of extralinguistic variation in language \citep{Wieling.2015}. We present what we believe to be the first attempt to incorporate geolinguistic knowledge into PLMs in a pretraining step, i.e., \emph{before task-specific fine-tuning}, making it possible to exploit it in any task for which it is expected to be useful. Specifically, we 
conduct an intermediate training step \cite{Glavas.2021} in the form of task-agnostic \textit{adaptation} -- dubbed  \textit{geoadaptation} -- that couples language modeling with predicting the geographic location (i.e., longitude and latitude) on geolocated texts. We choose adaptation as opposed to pretraining from scratch for three reasons: (i) intermediate training on language modeling (i.e., adaptation) before task-specific fine-tuning has proved beneficial for many NLP tasks \citep{Gururangan.2020},
(ii) adaptation has a lower computational cost
than pretraining \citep{Strubell.2019}, and (iii) PLMs encoding general-purpose linguistic knowledge are readily
available \cite{Wolf.2020}.\footnote{Notice that for the language areas we consider, there is
currently also not enough geotagged data that would allow us
to geographically pretrain models from scratch.} The specific method we introduce for geoadaptation combines language modeling with token-level geolocation prediction via multi-task learning, with task weights based on the homoscedastic uncertainties of the task losses \citep{Kendall.2018}.

We evaluate our geoadaptation framework on three groups of closely related languages, each with a corresponding PLM: (i) the German dialects spoken in Austria, Germany, and Switzerland (AGS) and GermanBERT, (ii) Bosnian-Croatian-Montenegrin-Serbian (BCMS) and BERTi\'{c}, and (iii)  Danish, Norwegian, and Swedish (DNS) and ScandiBERT. These groups exhibit strong geographic differences, providing an ideal testbed for geoadaptation.\footnote{Our focus on AGS, BCMS, and DNS also contributes to the recent call for more work on languages other than English in NLP \citep{joshi2020state,razumovskaia2021crossing}.} We further test geoadaptation at scale by adapting mBERT, a multilingual PLM, on the union of AGS, BCMS, and DNS.

We evaluate the effectiveness of geoadaptation on five downstream tasks expected to benefit from geolinguistic knowledge: (i) fine-tuned (i.e., supervised) geolocation prediction, (ii) zero-shot (i.e., unsupervised) geolocation prediction, (iii) fine-tuned language identification, (iv) zero-shot language identification, and (v) zero-shot prediction of dialect features. Geoadaptation leads to consistent performance gains compared to baseline models adapted on the same data using only language modeling, with particularly striking improvements on all zero-shot tasks. On two popular benchmarks for geolocation prediction and language identification, geoadaptation establishes a new state of the art. Furthermore, we show that geoadaptation \textit{geographically retrofits} the representation space of the PLMs. Overall, we see our study as an exciting step towards grounding PLMs in geography.\footnote{We make our code available at \url{https://github.com/valentinhofmann/geoadaptation}.}
\section{Related Work} \label{sec:related-work}

\textbf{Adaptation of PLMs.}
Continued language modeling training (i.e.,
adaptation) on data that comes from a similar distribution as the task-specific target data has been shown to improve the performance of PLMs for many NLP tasks \cite{glavavs2020xhate,Gururangan.2020} as well as in various language \cite{pfeiffer2020mad,parovic2022bad} and domain adaptation scenarios \cite{chronopoulou2021efficient,hung2022ds}. 
Adaptation can be seen as a special case of \emph{intermediate training}, which aims at improving the target-task performance of PLMs
by carrying out additional training between pretraining and fine-tuning \citep{Phang.2018,Vu.2020,Glavas.2021}. Intermediate training has also been conducted in a multi-task fashion, encompassing two or more training objectives \citep{Liu.2019c, Aghajanyan.2021}. 
Our work differs from these efforts in that it injects geolinguistic knowledge -- a type of extralinguistic knowledge -- into PLMs.

\vspace{1.4mm}
\noindent \textbf{Extralinguistic knowledge.} Leaving aside the large body of work on injecting visual \cite[e.g.,][]{bugliarello2022iglue} and structured knowledge \cite[e.g.,][]{lauscher2020common} into PLMs, a few studies have examined 
the interplay of PLM adaptation and extralinguistic factors \citep{Luu.2021, Rottger.2021}. However, they focus on \textit{time} and adapt PLMs to \emph{individual} extralinguistic contexts (i.e., time points). In contrast, we inject \textit{geographic} information from \emph{all} contexts into the PLM, forcing it to learn links between linguistic variability and a language-external variable -- in our case, geography. This is fundamentally different from adapting the PLM only to certain realizations of the language-external variable.

Most other studies introduce the extralinguistic information during task-specific fine-tuning \citep{Dhingra.2021, Hofmann.2021c, Karpov.2021, Kulkarni.2021,Rosin.2022b}. In contrast, we leverage geographic information only in the task-agnostic adaptation step. In task fine-tuning, the geoadapted PLM does not require any extralinguistic signal and is fine-tuned in the same manner as standard PLMs.

\vspace{1.4mm}
\noindent \textbf{Geography in NLP.}
We also build upon the long line of NLP research on geography, which roughly falls into two camps. On the one hand, many studies model geographically-conditioned 
differences in language, pointing to \emph{lexical variation} as the most conspicuous
manifestation \citep{Eisenstein.2010, Eisenstein.2011, Doyle.2014, Eisenstein.2014, Huang.2016, Hovy.2018, Hovy.2020},
although phonological \citep{Hulden.2011, Blodgett.2016}, syntactic \citep{Dunn.2019, Demszky.2021}, and semantic properties \citep{Bamman.2014, Kulkarni.2016} have been shown to exhibit geographic variation as well. On the other hand, there exists a large body of work on predicting geographic location from text, 
a task referred to as \emph{geolocation prediction} \citep{Rahimi.2015, Rahimi.2015b, Rahimi.2017, Salehi.2017, Rahimi.2018, Scherrer.2020, Scherrer.2021}. To the best of our knowledge, we are the first to geographically adapt PLMs in a task-agnostic fashion, making them more effective for any downstream task for which geolinguistic knowledge is relevant, from geolocation prediction to dialect-related tasks and language identification.
\section{Geoadaptation} \label{sec:method}

Let $\mathcal{D}$ be a geotagged dataset
consisting of sequences of tokens $X = (x_1, \dots, x_n)$ 
and corresponding geotags $T = (t_{\lon}, t_{\lat})$, where $t_{\lon}$
and $t_{\lat}$ denote the geographic longitude and latitude.
We want to adapt a PLM in such a way that 
it encodes the geographically-conditioned linguistic
variability in $\mathcal{D}$. Acknowledging the prominence of lexical variation among geographic differences in language (see \S\ref{sec:related-work}),
we accomplish this by
combining masked language modeling  (i.e., the pretraining objective) with token-level geolocation prediction in
a multi-task setup that pushes the PLM to learn associations between linguistic phenomena and geolocations \emph{on the lexical level}.\footnote{In this work, we focus on PLMs pretrained via masked language modeling. However, geoadaptation can in principle also be applied to autoregressive PLMs.}

\vspace{1.4mm}
\noindent\textbf{Masked language modeling.} We replace some tokens $x_i$ in $X$ with 
masked tokens $\tilde{x}_i$. Following \citet{Devlin.2019},
$\tilde{x}_i$ can be a special mask token (\texttt{[MASK]}), a random vocabulary token, or the original token itself. 
$X$ is fed into the PLM, which outputs a sequence of representations $E = (\mathbf{e}(x_1), \dots, 
\mathbf{e}(x_n))$. 
The representations of the masked tokens $\mathbf{e}(\tilde{x}_i)$ are then fed into a classification head. We compute the masked language modeling loss $\mathcal{L}_{\mlm}$ as the negative log-likelihood of the probability assigned to the true token.

\begin{figure*}[t!]
        \centering      
        \begin{subfigure}[b]{0.32\textwidth}  
            \includegraphics[trim={1.5cm 1cm 1.5cm 1cm},clip,width=\textwidth]{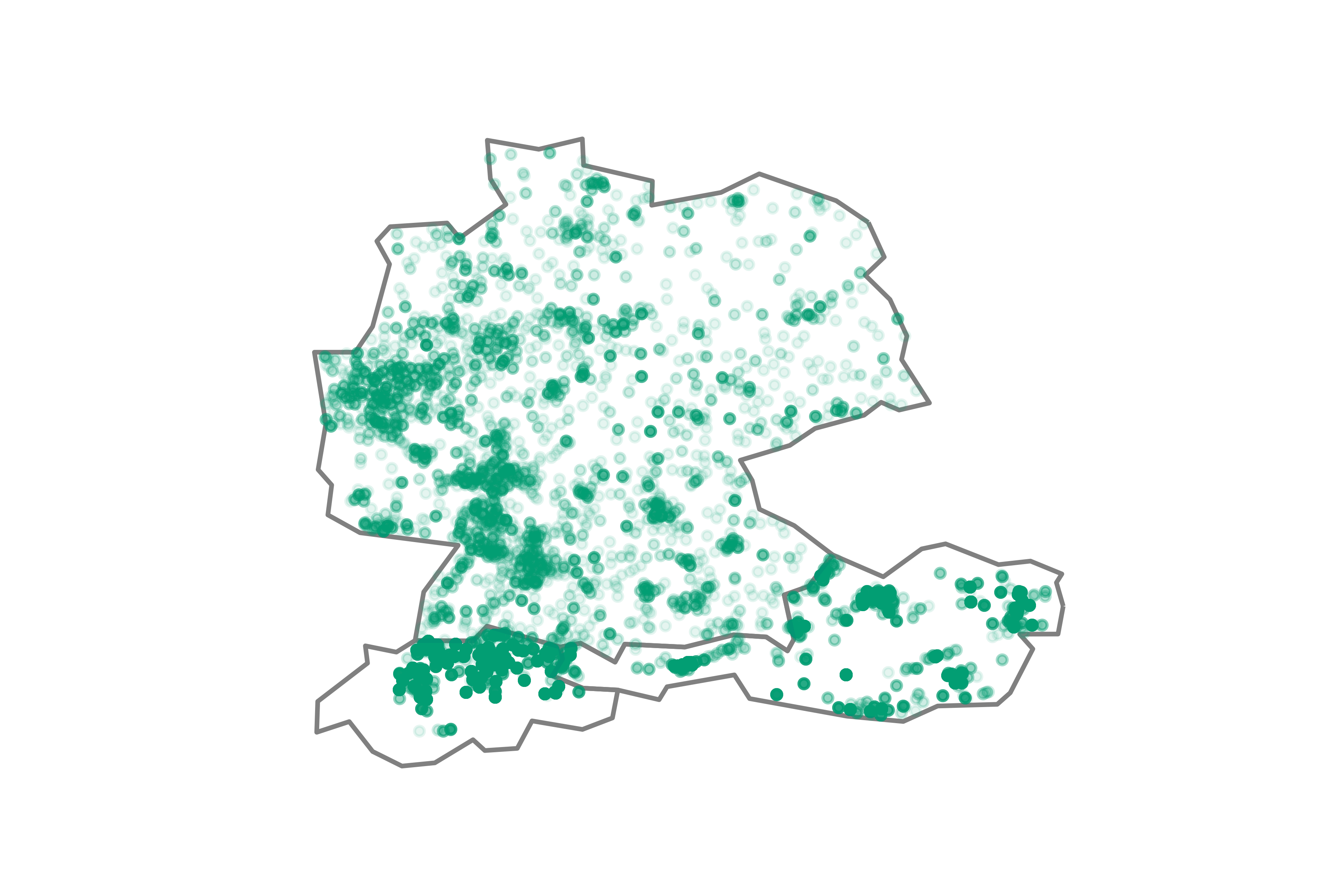}
            \caption[]%
            {{\small AGS}}    
        \end{subfigure}   
                \begin{subfigure}[b]{0.32\textwidth}  
            \includegraphics[trim={1.5cm 1cm 1.5cm 1cm},clip,width=\textwidth]{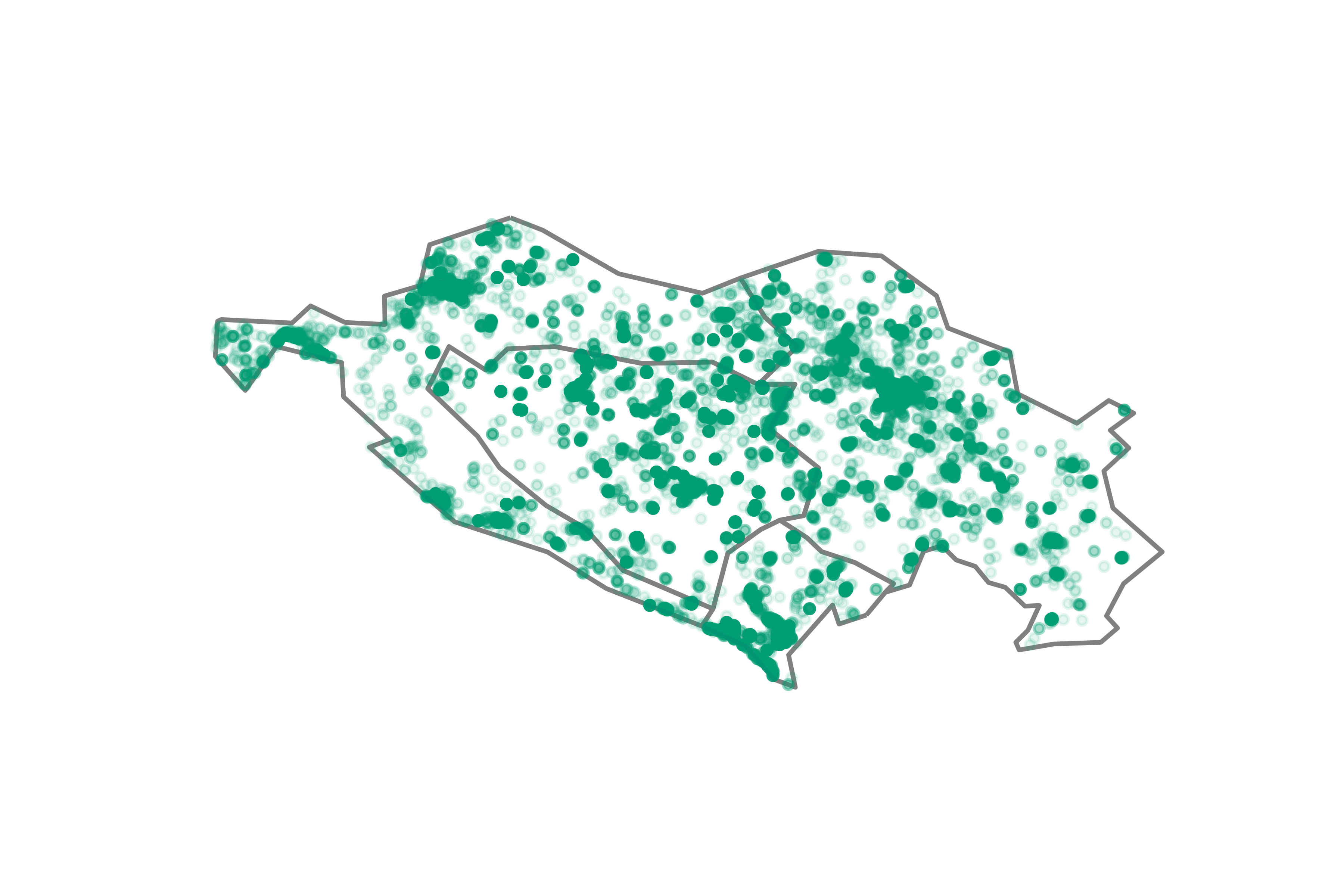}
            \caption[]%
            {{\small BCMS}}    
        \end{subfigure}     
        \begin{subfigure}[b]{0.32\textwidth}   
            \includegraphics[trim={1.5cm 1cm 1.5cm 1cm},clip,width=\textwidth]{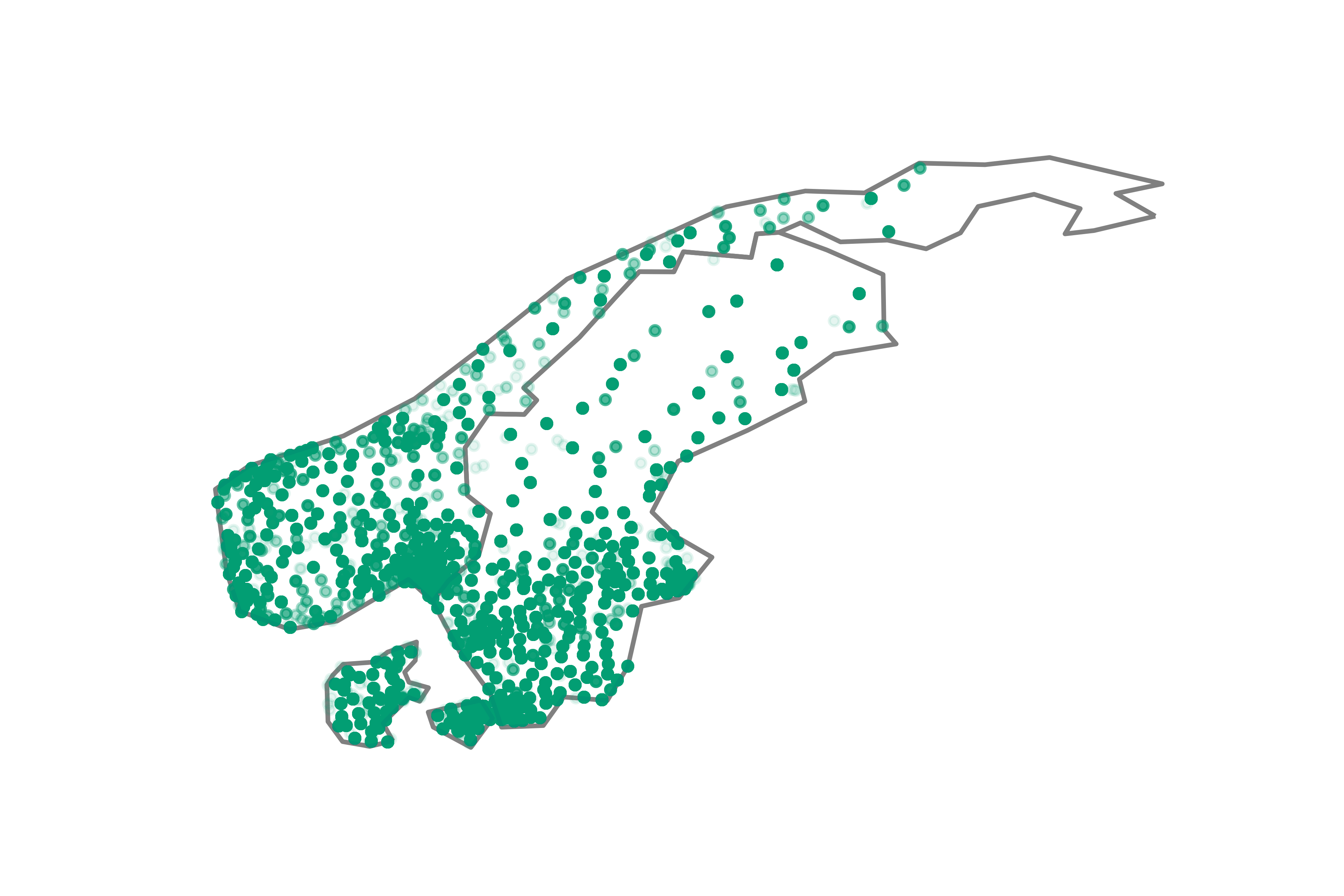}
            \caption[]%
            {{\small DNS}}   
        \end{subfigure}
        \caption[]{ Geographic distribution of the data for AGS (left), BCMS (middle), and DNS (right). Each point represents a Jodel post (AGS) or tweet (BCMS, DNS). Point density correlates with population density, with the densest areas corresponding to urban centers. For DNS, we exclude the Svalbard islands, which do not have any points. }
        \label{fig:geodistribution}
        \squeezeup
\end{figure*}

\vspace{1.4mm}
\noindent\textbf{Geolocation prediction.}
We additionally feed the vectors of masked tokens 
$\mathbf{e}(\tilde{x}_i)$ into a
feed-forward regression head that predicts two real-values: longitude and latitude. The geolocation prediction loss
$\mathcal{L}_{\geo}$ is the mean of the absolute prediction
errors for longitude and latitude. Note that the gold geolocation is the same for all masked tokens from the same input sequence. We inject geographic information at the token level because lexical variation represents the most prominent type of geographic language variation (see \S\ref{sec:related-work}).

\vspace{1.4mm}
\noindent\textbf{Composite multi-task loss.} 
We experiment with two different ways to compute the composite multi-task loss $\mathcal{L}_{\mt}$. First, we
straightforwardly sum the two task-specific losses: $\mathcal{L}_{\mt} = \mathcal{L}_{\mlm} + \mathcal{L}_{\geo}$. In multi-task training, however, a simple sum of the losses can be a suboptimal choice, especially if the losses are not of the same order of magnitude. In our case, $\mathcal{L}_{\mlm}$ and $\mathcal{L}_{\geo}$ are measured on different scales and relatively small values of $\mathcal{L}_{\geo}$ may still be multiples of relatively large values of $\mathcal{L}_{\mlm}$ (or vice versa). In a similar vein, the model might be more confident about
one task than about the other (e.g., associating contextual
token representations with geolocations may be easier than language modeling, i.e., predicting the correct token). To account for both factors, as a second method we compute the weights with which $\mathcal{L}_{\geo}$ and $\mathcal{L}_{\mlm}$ contribute to the joint loss based on their homoscedastic (i.e., task-dependent) uncertainties $\sigma_{\mlm}$
and $\sigma_{\geo}$ \citep{Kendall.2017}. $\sigma_{\mlm}$
and $\sigma_{\geo}$ are learned as part of the model training. The dynamic weighting ensures that the objectives are given equal importance with respect to the overall optimization.
Defining $l \in \{\mlm, \geo\}$,
we follow \citet{Kendall.2018} and replace $\mathcal{L}_l$ with:

\vspace{-0.5em}
\begin{equation}\label{eq:uncertainty}
\tilde{\mathcal{L}}_l = \frac{1}{2 \sigma_l^2}\mathcal{L}_l + \log \sigma_l.
\end{equation}
\vspace{-0.5em}

\noindent Equation \ref{eq:uncertainty} holds 
for both regression (e.g., mean absolute error as for  $\mathcal{L}_{\geo}$) and classification losses (e.g., categorical cross-entropy as for $\mathcal{L}_{\mlm}$) and can be derived from their Bayesian formulations \citep{Kendall.2018}. Notice that $\tilde{\mathcal{L}}_l$ is smoothly differentiable 
and well-formed: $\log \sigma_l$ ensures that the task weight $1/ \sigma_l^2$ does not converge to zero (or $\sigma_l^{2}$ diverges to infinity), which is the trivial solution to minimizing $1/{(2 \sigma_l^2)}\mathcal{L}_l$. For numerical stability, we set $\eta_l = 2 \log \sigma_l$ and compute $\tilde{\mathcal{L}}_l$ as:

\vspace{-0.5em}
\begin{equation}
\label{eq:eta_weight}
\tilde{\mathcal{L}}_l = \frac{1}{2} ( e^{-\eta_l}\mathcal{L}_l + \eta_l).
\end{equation}
\vspace{-0.5em}

\noindent The final multi-task loss is the sum of the two uncertainty losses: $\tilde{\mathcal{L}}_{\mt} = \tilde{\mathcal{L}}_{\mlm} + \tilde{\mathcal{L}}_{\geo}$.
\section{Experimental Setup}
\label{sec:exp}

\textbf{Models.} We examine four PLMs in this paper. For AGS, we use GermanBERT, a German BERT \citep{Devlin.2019} model.\footnote{\url{https://huggingface.co/dbmdz/bert-base-german-cased}} For BCMS, we use BERTi\'{c} \citep{Ljubevsic.2021}, a BCMS ELECTRA \citep{Clark.2020} model.\footnote{\url{https://huggingface.co/classla/bcms-bertic}} 
We specifically use the generator, i.e., a BERT model. 
For DNS, we resort to ScandiBERT, an XLM-Roberta \citep{Conneau.2020} model pretrained on corpora from five Scandinavian languages.\footnote{\url{https://huggingface.co/vesteinn/ScandiBERT}} Since we are interested to see whether geoadaptation can be expanded to a larger geographical area (e.g., an entire continent), we also geoadapt mBERT, a multilingual 
BERT \citep{Devlin.2019} model, on the union of the AGS, BCMS, and DNS areas.\footnote{\url{https://huggingface.co/bert-base-multilingual-cased}} We refer to this setting as EUR.

\vspace{1.4mm}
\noindent
\textbf{Data.} We start with a general overview of the data used for the experiments. Details about data splits are provided when describing the setup for geoadaptation as well as the evaluation tasks. Figure~\ref{fig:geodistribution} shows the geographic distribution of the data. Tables~\ref{tab:data-stats-1} and \ref{tab:data-stats-2} list summary statistics.

For AGS, we use the German data of the 2021 VarDial shared task on geolocation prediction \citep{Chakravarthi.2021}, which consist of geotagged Jodel posts from the AGS area. We merge the Austrian/German and Swiss portions of the data. For BCMS, 
we use the BCMS data of the 2021 VarDial shared task on geolocation prediction \citep{Chakravarthi.2021}, which consist 
of geotagged tweets from the BCMS area. To remedy  the sparsity of the data for
some regions, we retrieve an  additional set of geotagged tweets from the BCMS area posted between 2008 and 2021 using the Twitter API,
ensuring that there is no overlap with the VarDial data. For evaluation, we additionally draw upon SETimes, a news dataset for discriminating between Bosnian, Croatian, and Serbian \citep{Rupnik.2023}. For DNS, we use geotagged tweets from the Nordic Tweet Stream \citep{Laitinen.2018}, confining geotags to the DNS area.\footnote{For the sake of simplicity, in the following we will refer to both Jodel posts and tweets as \textit{posts}.} For evaluation, we additionally use the DNS portion of NordicDSL, a dataset of Wikipedia snippets for discriminating between Nordic languages \citep{Haas.2021}. For EUR, we mix the AGS, BCMS, and DNS data.

\begin{table*} [t!]
	\footnotesize
  \addtolength{\tabcolsep}{-1.5pt}
\centering
\begin{tabular}{lrrrrrrrrrrr}
\toprule
  & & \multicolumn{3}{c}{\ftg} & & \multicolumn{3}{c}{\ftl} & & \multicolumn{2}{c}{\zsd}\\ 
\cmidrule(lr){3-5}\cmidrule(lr){7-9}\cmidrule(lr){11-12}
Language & Adaptation & Train & Dev & Test & \zsg & Train & Dev & Test & \zsl & Phon & Lex\\ \midrule
AGS & 15,000 & 343,748 & 31,538 & 33,953 & 1,600 & 45,000 & 4,500 & 4,500 & --- & --- & ---\\
BCMS & 80,000 & 353,953 & 38,013 & 4,189 & 1,400 & 60,000 & 6,000 & 6,000  & 6,000 & 640 & 610\\
DNS & 300,000 & 150,000 & 75,000 & 75,000 & 3,900 & 45,000  & 4,500 & 4,500 & 4,500 & --- & ---\\
EUR & 50,000 & 100,000 & 10,000 & 10,000 & 4,500 & 100,000 & 10,000  &  10,000 & --- & --- & ---\\
\bottomrule
\end{tabular}
\caption{Data statistics. The table provides the number of Jodel posts (AGS), tweets (BCMS, DNS), or both (EUR) used for (geo-)adaptation and the five evaluation tasks (\ftg, \zsg, \ftl, \zsl, \zsd). There is no overlap between the Jodel posts/tweets used for (geo-)adaptation and the ones used for evaluation. The \ftg\ splits for AGS and BCMS are the original VarDial \citep{Chakravarthi.2021} splits.}  \label{tab:data-stats-1}
\squeezeup
\end{table*}

\begin{table} [t!]
	\footnotesize
\addtolength{\tabcolsep}{-1.5pt}
\centering
\begin{tabular}{lrrrr}
\toprule
  & \multicolumn{3}{c}{\ftl} & \\ 
\cmidrule(lr){2-4}
Language & Train & Dev & Test & \zsl \\ \midrule
BCMS & 7,374 & 963 & 921 & 921 \\
DNS & 22,796 & 5,699 & 1,497 & 1,497 \\
\bottomrule
\end{tabular} \caption{Out-of-domain data statistics. The table provides the number of news articles (BCMS) and Wikipedia snippets (DNS) used for out-of-domain \ftl\ and \zsl. The \ftl\ splits are the original SETimes \citep{Rupnik.2023} and NordicDSL \citep{Haas.2021} splits.}  \label{tab:data-stats-2}
\squeezeup
\end{table}

\vspace{1.4mm}
\noindent
\textbf{Geoadaptation.} For AGS, we create a balanced subset of the VarDial train posts (5,000 per country).\footnote{In preliminary experiments, we found that geographically-balanced sampling is beneficial for geoadaptation.} For BCMS, we draw upon the union of the VarDial train posts and the newly-collected posts to create 
a balanced subset (20,000 per country). For DNS, we similarly create a balanced subset of the posts (100,000 per country). 
For EUR, we sample balanced subsets of the AGS, BCMS, and DNS geoadaptation data (5,000 per country).
Using these four datasets, we adapt the PLMs
via the proposed multi-task learning approach (see~\S\ref{sec:method}). We geoadapt 
the PLMs for 25 epochs and save the model snaphots after each epoch. To track progress, 
we measure perplexity and token-level median distance on the VarDial development sets for AGS and BCMS, a separate set of 75,000 posts for DNS, and a separate set of 10,000 posts for EUR.

\vspace{1.4mm}
\noindent
\textbf{Evaluation tasks.} Inspired by existing NLP research on
geography (see \S\ref{sec:related-work}), we evaluate the geoadapted PLMs on five tasks that probe different aspects 
of the learned associations between linguistic phenomena and geography.  

\vspace{1.4mm}
\noindent
\textit{Fine-tuned geolocation prediction (\ftg).} We fine-tune the geoadapted PLMs for geolocation prediction. For AGS and BCMS, we use the train, dev, and test splits from VarDial. For DNS, we create separate sets of train, dev, and test posts; we do the same for EUR, drawing train, dev, and test posts from the union of the AGS, BCMS, and DNS data (see Table~\ref{tab:data-stats-1}). We make sure that there is no overlap between the geoadaptation posts and dev and test posts of any of the downstream evaluation tasks. Following prior work by \citet{Scherrer.2021}, we cast geolocation prediction as a multi-class classification task: we first map all geolocations in the train sets into $k$ clusters using
$k$-means and assign each geotagged post to its closest cluster.\footnote{We standardize longitude and latitude values and use the Euclidean distance as the clustering metric. Following \citet{Scherrer.2021}, we choose $k=75$.} Concretely, we pass the contextualized vector of the \texttt{[CLS]} token to a single-layer softmax classifier that outputs probability distributions over the $k$ geographic clusters.

In line with prior work, we use the median of the Euclidean distance between the predicted and true geolocation as the evaluation metric. Note that \ftg\ is \emph{different} from geolocation prediction in geoadaptation (see \S\ref{sec:method}): there, we (i) cast geolocation prediction as a regression task (i.e., predict the exact longitude and latitude) and (ii) predict the geolocation from the masked tokens, rather than the representation of the whole post.

\vspace{1.4mm}
\noindent
\textit{Zero-shot geolocation prediction (\zsg).} 
Given the central objective of geoadaptation (i.e., to induce mappings between linguistic variation and geography), we next test if the geoadapted models can predict geographic information from text without any fine-tuning. To this end, we directly probe the PLMs for geolinguistic associations: with the help of prompts, we ask the PLMs to generate the correct toponym corresponding to a post's geolocation using their language modeling head, which has not been trained on geolocation prediction in any way (see~\S\ref{sec:method}). 
We do this on the most fine-grained geographic resolution possible, i.e., cities for BCMS/DNS and states for AGS.\footnote{Most posts in the AGS data come from rural areas.} For EUR, we draw upon the union of AGS, BCMS, and DNS, resulting in a mix of cities and states.

To create the data for \zsg, we start by reverse-geocoding all posts and then select cities/states that 
contain at least 100 posts and have names existing in the PLM vocabulary. We randomly sample 100 posts from each of these cities/states (AGS: \ul{\textit{Bayern}},
\ul{\textit{Bern}},
\ul{\textit{Brandenburg}},
\ul{\textit{Bremen}},
\ul{\textit{Hessen}},
\ul{\textit{K\"arnten}},
\ul{\textit{Luzern}},
\ul{\textit{Niedersachsen}},
\ul{\textit{Ober\"osterreich}},
\textit{Saarland},
\ul{\textit{Sachsen}},
\ul{\textit{Salzburg}},
\ul{\textit{Steiermark}},
\ul{\textit{Th\"uringen}},
\ul{\textit{Tirol}},
\ul{\textit{Z\"urich}}; BCMS: \ul{\textit{Bar}}, \ul{\textit{Beograd}}, \textit{Bor}, \ul{\textit{Dubrovnik}}, \textit{Kragujevac}, \textit{Ni\v{s}}, \ul{\textit{Podgorica}}, 
\textit{Pula}, 
\ul{\textit{Rijeka}}, \ul{\textit{Sarajevo}}, \ul{\textit{Split}}, \textit{Tuzla}, 
\ul{\textit{Zagreb}}, \textit{Zenica}; DNS: \ul{\textit{Aalborg}},
\ul{\textit{Aarhus}},
\textit{Arendal},
\ul{\textit{Bergen}},
\textit{Drammen},
\ul{\textit{Fredrikstad}},
\ul{\textit{G\"oteborg}},
\textit{Halmstad},
\textit{Haugesund},
\textit{Helsingborg},
\ul{\textit{Kalmar}},
\textit{Karlstad},
\ul{\textit{Kristiansand}},
\ul{\textit{K{\o}benhavn}},
\textit{Linköping},
\textit{Luleå},
\ul{\textit{Lund}},
\ul{\textit{Moss}},
\ul{\textit{Nora}},
\textit{Norrköping},
\ul{\textit{Odense}},
\ul{\textit{Oslo}},
\textit{Porsgrunn},
\ul{\textit{Roskilde}},
\ul{\textit{Sala}},
\textit{Sandefjord},
\textit{Sarpsborg},
\textit{Skien},
\ul{\textit{Stavanger}},
\ul{\textit{Stockholm}},
\textit{Södertälje},
\ul{\textit{Troms{\o}}},
\ul{\textit{Trondheim}},
\textit{T{\o}nsberg},
\textit{Uddevalla},
\ul{\textit{Ume\aa}},
\ul{\textit{Uppsala}},
\textit{Ålesund},
\ul{\textit{\"Orebro}}; EUR: 45 underlined cities/states above, which are in the mBERT vocabulary).

For zero-shot prediction, we append prompts with the meaning `This is [MASK]' to the post (AGS: \textit{Das ist [MASK]}; BCMS: \textit{To je [MASK]}; DNS: \textit{Dette er [MASK]}).\footnote{We experimented with other prompts (e.g., `This is in [MASK]') and obtained similar results.} For EUR, we just append \textit{[MASK]} to the post. We pass the whole sequence to the PLM and forward the output representation of the \texttt{[MASK]} token into the language modeling head. Following common practice \citep{Xiong.2020}, 
we restrict the output vocabulary to the set of candidate labels, i.e., we select 
the city or state name with the highest logit. We measure the performance in terms of accuracy.

\vspace{1.4mm}
\noindent
\textit{Fine-tuned language identification (\ftl).} Next, we consider language identification, a task of great importance for many applications that is particularly challenging in the case of closely related languages \citep{Zampieri.2014, Haas.2021}. While arguably less directly tied to geography than geolocation prediction, we believe that language identification should also benefit from geoadaptation since one or (in the case of multilingual communities) few languages are used at any given location -- having knowledge about geolinguistic variation should thus make it easier to distinguish different languages.

We start by \emph{fine-tuning} the PLMs for language identification. For AGS, BCMS, and DNS, we reuse the respective \ftg\ datasets and
sample 15,000 train, 1,500 dev, and 1,500 test posts per language (determined based on their geolocation). For EUR, we reuse the exact \ftg\ train, dev, and test split. To test how well the effects of geoadaptation generalize to out-of-domain data, we also fine-tune BERTi\'{c} on SETimes (i.e., news articles) and ScandiBERT on NordicDSL (i.e., Wikipedia snippets). In terms of modeling, we formulate language identification as a multi-class classification task, with three classes for AGS/DNS, four classes for BCMS, and 10 classes for EUR. We again pass the contextualized vector of the \texttt{[CLS]} token to a single-layer softmax classifier that outputs probability distributions over the languages. We measure the performance in terms of accuracy.

\vspace{1.4mm}
\noindent
\textit{Zero-shot language identification (\zsl).} Similarly to geolocation prediction, we are interested to see how well the geoadapted PLMs can identify the language of a text without fine-tuning. We reuse the \ftl\ test sets for this task. The setup follows \zsg, i.e., we append the same prompts to the posts, pass the full sequences through the PLMs, and feed the output representations of the \texttt{[MASK]} token into the language modeling head. However, instead of city/state names, we now consider language names, specifically \textit{bosanski} (`Bosnian'), \textit{crnogorski} (`Montenegrin'), \textit{hrvatski} (`Croatian'), and \textit{srpski} (`Serbian') in the case of BCMS, and \textit{dansk} (`Danish'), \textit{norsk} (`Norwegian'), and \textit{svensk} (`Swedish') in the case of DNS.\footnote{We do not conduct \zsl\ for AGS and EUR since the names of the German dialects (e.g., \textit{Schweizerdeutsch}) are not in the GermanBERT and mBERT vocabularies.} We select the language name with the highest logit and measure the performance in terms of accuracy.

\begin{table*} [t!]
\addtolength{\tabcolsep}{-2pt}
\footnotesize
\centering
\begin{tabular}{lrrrrrrrrrrrr}
\toprule
& \multicolumn{8}{c}{\ftg\ $\downarrow$}\\ 
\cmidrule(lr){2-9}
& \multicolumn{2}{c}{AGS} & \multicolumn{2}{c}{BCMS} & \multicolumn{2}{c}{DNS} & \multicolumn{2}{c}{EUR} & \multicolumn{4}{c}{\zsg\ $\uparrow$}\\
\cmidrule(lr){2-3}\cmidrule(lr){4-5}\cmidrule(lr){6-7}\cmidrule(lr){8-9}\cmidrule(lr){10-13}
Method & Dev & Test & Dev & Test & Dev & Test & Dev & Test & AGS & BCMS & DNS & EUR\\
\midrule
SotA / Rand & --- & --- &   $^?$30.11 & $^?$15.49  & --- & --- &    --- & --- & $^\ddag$.071 & $^\ddag$.070 & $^\ddag$.026 & $^\ddag$.021 \\
\vada & $^\ddag$193.51 & $^\ddag$196.18 & $^\ddag$29.36 & $^\ddag$16.72 & $^\ddag$101.15 & $^\ddag$101.15  & $^\ddag$107.20 & $^\ddag$107.41 & $^\ddag$.142 & $^\ddag$.144 & $^\ddag$.106 & $^\ddag$.108 \\ \midrule
\gadas & $^\dag$\second{190.21} & \first{193.18} & $^\dag$\second{26.02} & $^\dag$\second{13.98} & $^\dag$\second{98.82} & $^\dag$\second{97.63} &  $^\dag$\second{98.00} & $^\dag$\second{101.76} & \second{.192} & $^\dag$\second{.287} & $^\dag$\second{.135} & $^\dag$\second{.159}\\
\gadaw & \first{189.06} & $^\dag$\second{194.85} & \first{23.90} & \first{12.13} & \first{95.80} & \first{97.06} &  \first{97.18} & \first{97.18} &\first{.193} &  \first{.319} & \first{.149} & \first{.191}\\
\bottomrule
\end{tabular}
\caption{Results on fine-tuned geolocation prediction (\ftg) and zero-shot geolocation prediction (\zsg). Measure for \ftg: median distance (in km); measure for \zsg: prediction accuracy. For \ftg\ and BCMS, the first row shows the current state-of-the-art performance \cite{Scherrer.2021}. For \zsg, the first row shows random performance. \textbf{Bold}: best score in each column; \underline{underline}: second best score. We highlight scores that are significantly ($p < .05$) worse than the best score with a $^\dag$ and scores that are significantly ($p < .05$) worse than the two best scores with a $^\ddag$. We indicate with a $^?$ scores for which we cannot test for statistical significance since we do not have access to the distribution of output predictions.}  
\label{tab:ftg}
\squeezeup
\end{table*}

\vspace{1.4mm}
\noindent
\textit{Zero-shot dialect feature prediction (\zsd).} The fifth evaluation tests whether geoadaptation 
increases the PLMs' awareness of dialectal variation. We only conduct this task for BCMS, which exhibits many well-documented dialectal variants that exist as tokens in the BERTi\'{c} vocabulary.

We consider two subtasks. In the first subtask (Phon), we test whether BERTi\'{c} can select the
correct variant for a phonological variable, specifically the reflex of the Old Slavic vowel \textit{\v{e}}. This feature exhibits 
geographic variation in BCMS: in the
(north-)west,
the reflexes \textit{ije} and \textit{je} are predominately used, whereas the
(south-)east  mostly uses \textit{e}
\citep{Ljubesic.2018}, e.g., \textit{l\underline{ije}po}
vs.\ \textit{l\underline{e}po} (`nice'). Drawing upon words for which both
  \textit{ije}/\textit{je} and \textit{e} variants exist in
  the BERTi\'{c} vocabulary, we filter out words
  that appear in fewer than 10
  posts in the merged VarDial dev and test data, resulting in a set of 64 words (i.e., 32 pairs). Subsequently, we randomly sample 10
  posts for each of the words. For the second subtask (Lex), we evaluate the recognition of lexical variation that is not tied to a phonological feature \citep{Alexander.2006}, e.g., \textit{porodica} vs.\ \textit{obitelj} (`family'). 
Based on a Croatian-Serbian comparative dictionary,\footnote{\url{https://hr.wiktionary.org/wiki/Razlikovni_rje\%C4\%8Dnik_hrvatskog_jezika_i_srpskog_jezika}} we select all pairs for which both words are in the BERTi\'{c}
vocabulary. We remove words that occur in fewer than 10 VarDial dev and test posts and sample 10 posts for each of the remaining 61 words.

For prediction, we mask out the phonological/lexical variant and follow the same approach as for \zsg\ and \zsl, with the difference that we restrict the vocabulary to the two relevant variants (e.g., \textit{porodica} vs.\ \textit{obitelj}). We measure the performance in terms of accuracy.

\vspace{1.4mm}
\noindent
\textbf{Model variants.}
We evaluate the two geoadaptation variants, minimizing the simple sum of $\mathcal{L}_{\mlm}$ and $\mathcal{L}_{\geo}$ (\gadas) and the weighted sum based on homoscedastic uncertainty (\gadaw). To quantify the effects of geoadaptation compared to standard adaptation, we adapt the PLMs on the same data using only $\mathcal{L}_{\mlm}$ as the primary baseline (\vada), i.e., the \vada\ models are adapted \textit{on the exact same text data} as \gadas\ and \gadaw, but using continued language modeling training \textit{without} geolocation prediction. Where possible (i.e., BCMS \ftg\ and out-of-domain BCMS \ftl), we compare against the current state-of-the-art (SotA) performances \cite{Scherrer.2021, Rupnik.2023} -- BERTi\'{c} fine-tuned on the train data. On the zero-shot tasks, we also report random performance (Rand). 

Language identification is a task that is not typically addressed using PLMs. Instead, most 
state-of-the-art systems are less expensive models trained on character $n$-grams \citep{Zampieri.2017,Haas.2021, Rupnik.2023}. To 
get a sense of whether PLMs in general and geoadapted PLMs in particular are competitive with such 
custom-built systems, we evaluate GlotLID \citep{Kargaran.2023}, a strong 
language identification tool based on FastText
\citep{Bojanowski.2017, Joulin.2017}, on \ftl. Since GlotLID was not specifically trained on the domains examined in \ftl, 
we also train new FastText models on the data used to fine-tune the PLMs.

\vspace{1.4mm}
\noindent
\textbf{Hyperparameters.}
For geoadaptation, we use a batch size of 32 (16 for mBERT) and perform grid search for the learning rate $r \in \{ \num{1e-5}, \num{3e-5} , \num{1e-4} \}$. We always geoadapt the PLMs for 25 epochs. For \ftg, we use a batch size of 32 (16 for mBERT) and perform grid search for the number of epochs $n \in \{1, \dots ,10\}$
and the learning rate $r \in \{ \num{1e-5}, \num{3e-5} , \num{1e-4} \}$. For \ftl, we use a batch size of 32 (16 for mBERT) and perform grid search for the number of epochs $n \in \{1, \dots ,5\}$
and the learning rate $r \in \{ \num{1e-5}, \num{3e-5} , \num{1e-4} \}$. For all training settings (geoadaptation, \ftg, \ftl)
we tune $r$ for \vada\ only and use the best configuration for \gadaw\ and \gadas. This means that the overall number of hyperparameter trials is 
three times larger for \vada\ than \gadaw\ and \gadas, i.e., we are giving a substantial advantage to the models that serve as a baseline.
We use Adam \citep{Kingma.2015} as the optimizer. 
All experiments are performed on a GeForce GTX 1080 Ti GPU (11GB). For the FastText models trained on \ftl, we perform grid search for the number of epochs $n \in \{5, 10, 15, 20,25\}$, 
the minimum length of included character $n$-grams $l_{\min} \in \{1, 2, 3\}$, and the maximum length of included 
character $n$-grams $l_{\max} \in \{4, 5, 6\}$.

\begin{figure*}[ht!]
\centering      
        \begin{subfigure}[b]{0.32\textwidth}  
          
            \includegraphics[width=\textwidth]{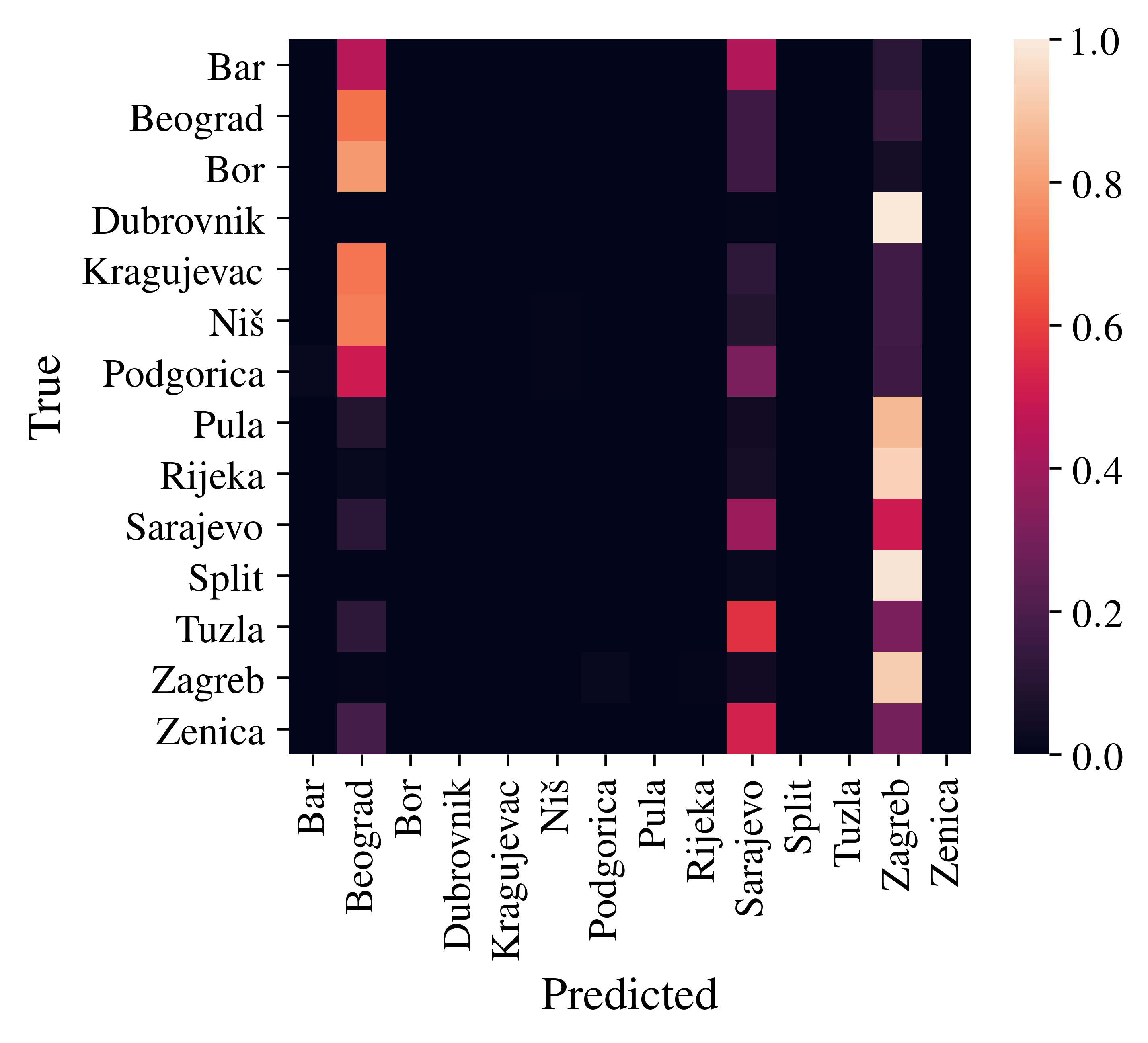}
            \caption[]%
            {{\small \vada}}    
        \end{subfigure}     
        \begin{subfigure}[b]{0.32\textwidth}   
         
            \includegraphics[width=\textwidth]{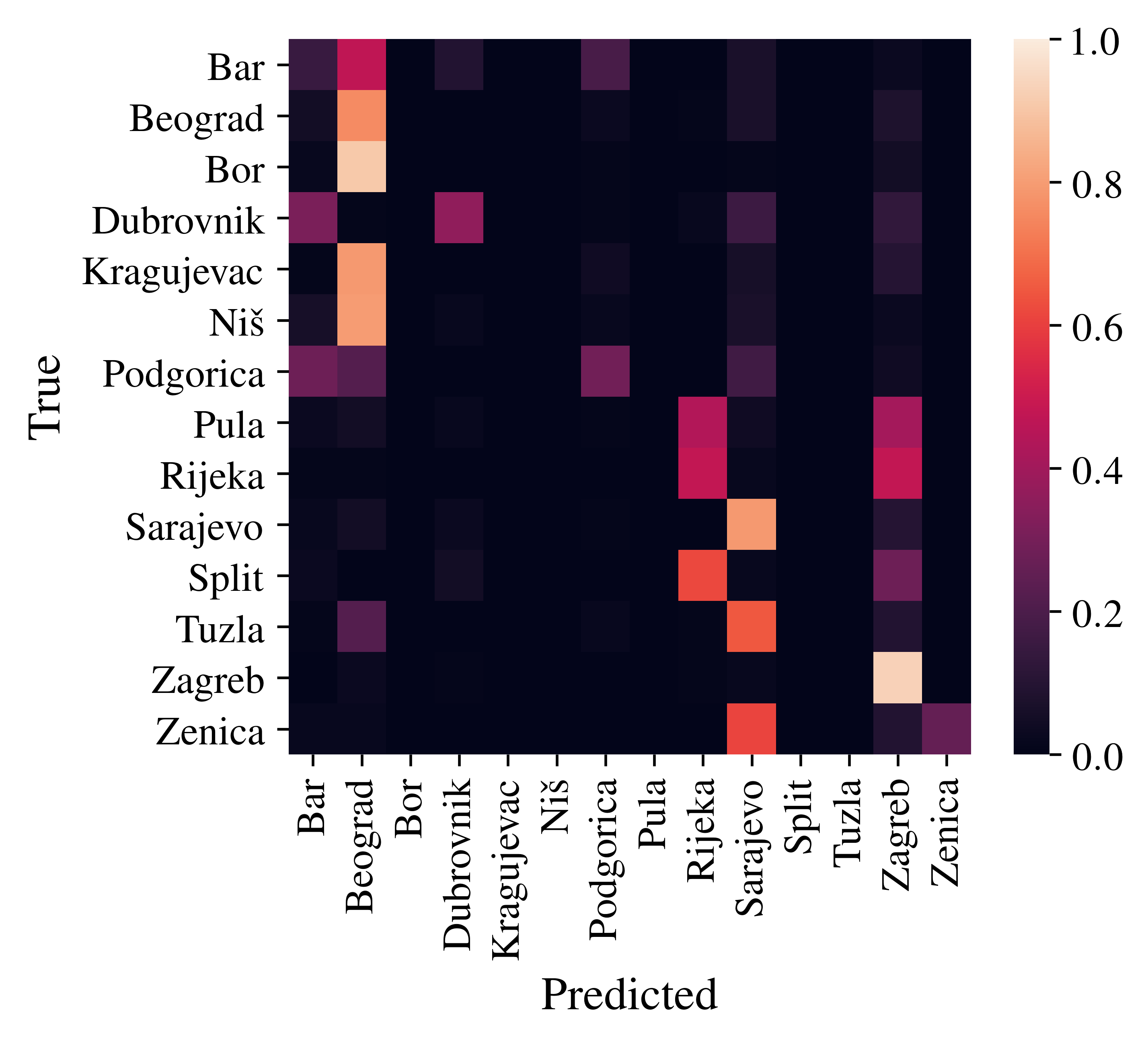}
            \caption[]%
            {{\small \gadas}}    
        \end{subfigure}
               \begin{subfigure}[b]{0.32\textwidth}   
         
            \includegraphics[width=\textwidth]{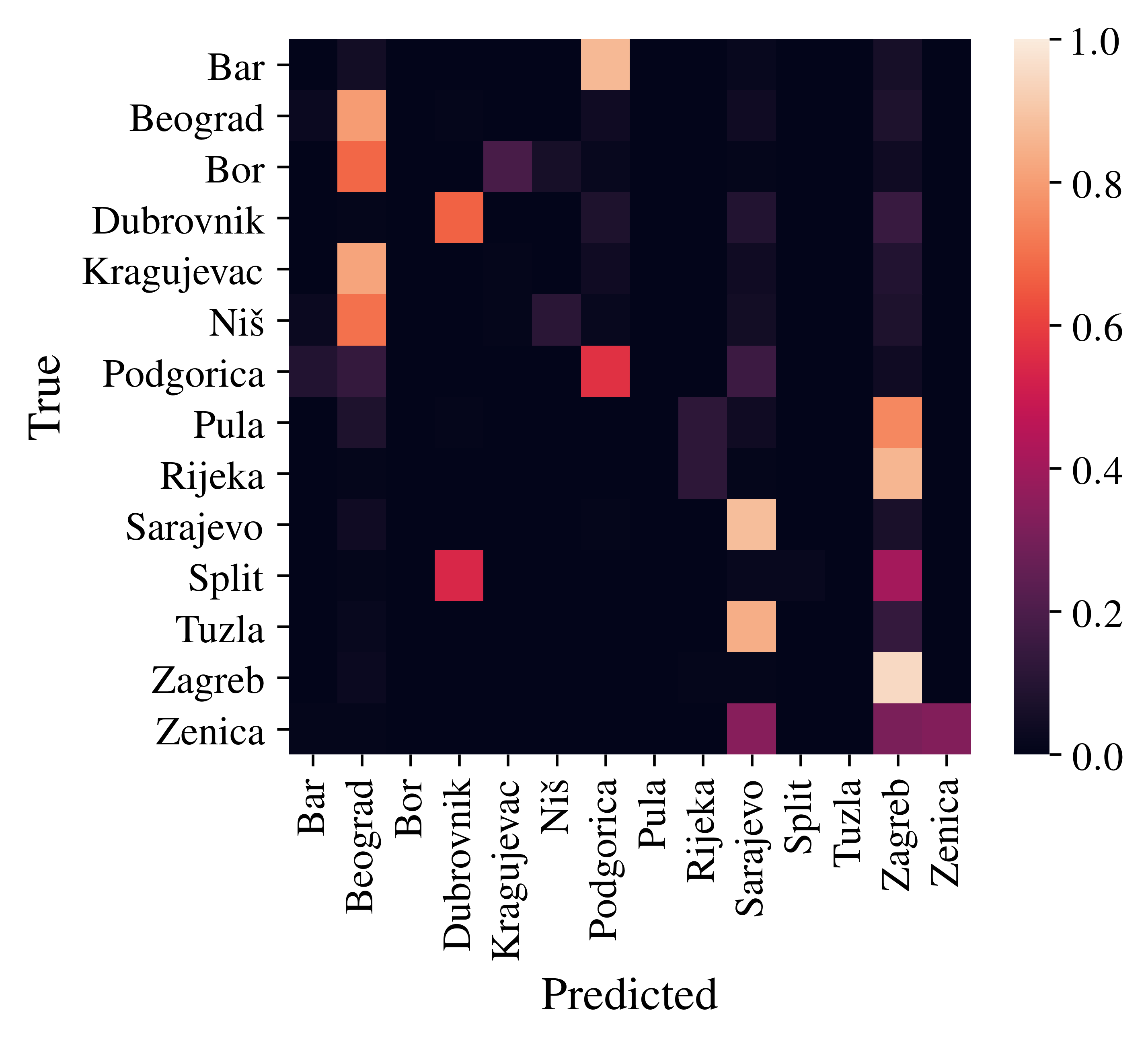}
            \caption[]%
            {{\small \gadaw}}    
        \end{subfigure}
        \caption[]{Confusion matrices for \vada\ (a), \gadas\ (b), and \gadaw\ (c) on \zsg\ (BCMS). While \vada\ always predicts one of the three most frequent city tokens (\textit{Beograd}, \textit{Sarajevo}, or \textit{Zagreb}), the predictions of \gadas\ and \gadaw\ are much more diverse and less tied to frequency.}
        \label{fig:confusion}
        \squeezeup
\end{figure*}

\section{Results and Analysis}
\label{sec:results}

Tables \ref{tab:ftg}, \ref{tab:ftl}, \ref{tab:ftl-ood}, and \ref{tab:zsd} compare the performance of the geoadapted PLMs against the baselines. To test for statistical significance of the performance differences, we use paired, two-sided Student's $t$-tests in the case of \ftg\ and McNemar's tests for binary data \citep{McNemar.1947} in the case of \zsg, \ftl, \zsl, and \zsd, as recommended by \citet{Dror.2018}. We correct the resulting $p$-values for each evaluation using the Holm-Bonferroni method \citep{Holm.1979}.

Overall, the geoadapted models consistently
and substantially outperform the baselines -- out of the 30 main evaluations, it is \textit{always} one of the two geoadapted models that achieves the best score, a result that is highly unlikely to occur by chance if there is no underlying performance difference between the geoadapted and non-geoadapted models.\footnote{Assuming equal underlying performance for \vada, \gadas, and \gadaw\ (and ignoring other baselines), the probability for this result is $p = (2 / 3)^{30} < 10^{-5}$.} Furthermore, in the two cases where we can directly compare to a prior state of the art, one or both geoadapted models outperform it. These findings strongly suggest that geoadaptation successfully induces associations between language variation and geographic location.

\vspace{1.4mm}
\noindent \textbf{Fine-tuned geolocation prediction.}
PLMs geoadapted with uncertainty weighting (\gadaw) predict the geolocation most precisely (see Table~\ref{tab:ftg}). On BCMS, \gadaw\ improves the previous state of the art -- achieved by a directly fine-tuned BERTi\'{c} model -- by 3.3 km on test and by over 6 km on dev. On EUR (arguably the most challenging setting), \gadaw\ improves upon \vada\ (i.e., a model adapted without geographic signal) by more than 10 km on both dev and test. \vada\ always performs worse than the two geoadapted models, despite the fact that task-specific fine-tuning likely compensates for some of the geographic knowledge \gadaw\, and \gadas\, obtain in geoadaptation. This shows that \emph{geoadaptation} drives the performance improvements, and that language modeling adaptation alone does not suffice.     
Loss weighting based on homoscedastic uncertainties seems beneficial for \ftg: while \gadas\ already outperforms the baselines, \gadaw\ in seven out of eight cases brings further significant gains.
We also observe that all models reach peak performance in the first few fine-tuning epochs (not shown), and that geoadaptation is useful even when the geoadaptation data are a subset of the fine-tuning data (as is the case for AGS).
This confirms that the performance gains come from the geoadaptation and are not merely the result of longer training on geolocation prediction.

\vspace{1.2mm}
\noindent \textbf{Zero-shot geolocation prediction.} 
In this task, the PLMs have to predict the \textit{token} of the correct toponym (i.e., city or state). Notice that the PLMs receive 
information about exact geolocations during geoadaptation and do not leverage
toponym tokens in any direct way. \zsg\ is thus an ideal litmus test as it shows how well the link between language variation and geography, injected into the PLMs via geoadaptation, generalizes. 
The results (see Table~\ref{tab:ftg}) strongly suggest that geoadaptation leads to such generalization: both geoadapted model variants bring massive and statistically significant gains in prediction accuracy over \vada\ (e.g., \gadaw\ vs.\ \vada: +17.5\% on BCMS, +8.3\% on EUR). As on FT-Geoloc, uncertainty weighting (\gadaw) overall outperforms simple loss summation (\gadas).

\begin{table*} [t!]
\addtolength{\tabcolsep}{-2pt}
\footnotesize
\centering
\begin{tabular}{lrrrr}
\toprule
& \multicolumn{4}{c}{\zsg\ $\uparrow$} \\ 
\cmidrule(lr){2-5}
Method & AGS  & BCMS & DNS & EUR \\
\midrule
\vada & $^\ddag$.156 \upbox{.014} & $^\ddag$.150 \upbox{.006} &  $^{\ddag*}$.131 \upbox{.025} & $^{\ddag*}$.139 \upbox{.031}\\ \midrule
\gadas & $^*$\first{.229} \upbox{.036} & $^*$\first{.386} \upbox{.099} &  \second{.147} \upbox{.012} & $^{\dag*}$\second{.195} \upbox{.036}\\
\gadaw & $^*$\first{.229} \upbox{.036} &  $^*$\second{.373} \upbox{.054} & \first{.152} \upbox{.003} & $^*$\first{.219} \upbox{.028}\\
\bottomrule
\end{tabular}
\caption{Results on zero-shot geolocation prediction (\zsg) with calibration \citep{Zhao.2021}. Measure: prediction accuracy. Besides the results, we give the changes compared to vanilla \zsg\ and indicate with a $^*$ if they are significant ($p < .05$). See Table~\ref{tab:ftg} for an explanation of the other symbols used in the table.}  
\label{tab:calibration}
\squeezeup
\end{table*}

Figure \ref{fig:confusion} shows the confusion matrices for the three methods on BCMS, offering further insights. \vada\ assigns most posts from a country
to the corresponding capital (e.g., posts from Croatian cities
to \textit{Zagreb}). These tokens are the most frequent ones
out of all considered cities, which seems to heavily affect \vada.
In contrast, predictions of \gadas\ and \gadaw\ are much more nuanced, i.e., more diverse and less tied to the frequency of the toponym tokens: the geoadapted models are not only able to correctly assign posts from smaller, less frequently mentioned cities (e.g., \textit{Dubrovnik}, \textit{Zenica}), but their errors also reflect regional linguistic consistency and geographic proximity. For example, \gadas\  predicts \textit{Rijeka} as the origin of many \textit{Pula} posts, and \textit{Bar} as the origin of many \textit{Dubrovnik} posts; similarly, \gadaw\  assigns posts from \textit{Split} to \textit{Dubrovnik} and posts from \textit{Bar} to \textit{Podgorica}.\footnote{Note that \textit{Bar} and \textit{Dubrovnik} are not in the same country.}

One common method to alleviate the impact of different prior probabilities in the zero-shot setting (a potential reason for the bad performance of \vada) is to \emph{calibrate} the PLM predictions \citep{Holtzman.2021, Zhao.2021}. Following \citet{Zhao.2021}, we measure the prior probabilities of all toponym tokens using a neutral prompt (specifically, `This is [MASK]' for AGS/BCMS/DNS and a \texttt{[MASK]} token for EUR) and repeat the \zsg\ evaluation, dividing the output probabilities by the prior probabilities (Table~\ref{tab:calibration}). We find that all models (both geoadapted and non-geoadapted) improve as a result of calibration, i.e., the output probabilities seem to be miscalibrated if not specifically adjusted by means of the prior probabilities. However, refuting the hypothesis that miscalibration causes the inferior performance of \vada, the average gain due to calibration is larger for the geoadapted models (\gadas: +4.8\%, \gadaw: +3.0\%) than for the non-geoadapted models (\vada: +1.9\%). This suggests that a miscalibration of the toponym probabilities -- rather than disproportionately affecting the non-geoadapted models -- generally impairs the geolinguistic capabilities of a PLM. The consequences of such an impairment seem to be the more detrimental the more profound the underlying geolinguistic knowledge.

Taken together, these observations indicate that
\gadas\ and \gadaw\ possess detailed knowledge
of geographic variation in language. Since geoadaptation provides no supervision in the form of toponym names, this implies an impressive generalization, i.e., the association of linguistic constructs to toponyms, with geolocations (specifically, scalar longitude-latitude pairs) as the intermediary signal driving the generalization.

\begin{table*} [t!]
\addtolength{\tabcolsep}{-2pt}
\footnotesize
\centering
\begin{tabular}{lrrrrrrrrrr}
\toprule
& \multicolumn{8}{c}{\ftl\ $\uparrow$}\\ 
\cmidrule(lr){2-9}
& \multicolumn{2}{c}{AGS} & \multicolumn{2}{c}{BCMS} & \multicolumn{2}{c}{DNS} & \multicolumn{2}{c}{EUR} & \multicolumn{2}{c}{\zsl\ $\uparrow$}\\
\cmidrule(lr){2-3}\cmidrule(lr){4-5}\cmidrule(lr){6-7}\cmidrule(lr){8-9}\cmidrule(lr){10-11}
Method & Dev & Test & Dev & Test & Dev & Test & Dev & Test & BCMS & DNS \\
\midrule
Rand & --- & --- &   --- & ---  & --- & --- &    --- & --- &   $^\ddag$.245 & $^\ddag$.339 \\
GlotLID & --- & --- & $^\ddag$.323 & $^\ddag$.316 & $^\ddag$.927 & $^\ddag$.931 & --- & --- & --- & --- \\
FastText & $^\ddag$.843 & $^\ddag$.840 & $^\ddag$.598 & $^\ddag$.588 & $^\ddag$.948 & $^\ddag$.959 & $^\ddag$.757 & $^\ddag$.762 & --- & --- \\
\vada & .851 & .855 & $^\ddag$.693 & $^\ddag$.694 & $^\ddag$.964 & $^\ddag$.966 & $^\ddag$.776 & $^\ddag$.777 & $^\ddag$.417 & $^\ddag$.885 \\ \midrule
\gadas & \first{.861} & \second{.856} & \second{.734} & \second{.726} & \second{.972} & \second{.975} & \second{.789} & $^\dag$\second{.786} & \first{.553} & $^\dag$\second{.896}\\
\gadaw & \first{.861} & \first{.858} & \first{.743} & \first{.734} & \first{.973} & \first{.976} & \first{.792} & \first{.796} & $^\dag$\second{.543} & \first{.927}\\
\bottomrule
\end{tabular}
\caption{Results on fine-tuned language identification (\ftl) and zero-shot language identification (\zsl). Measure: prediction accuracy. See Table~\ref{tab:ftg} for an explanation of the symbols used in the table.}  
\label{tab:ftl}
\squeezeup
\end{table*}

\begin{table} [t!]
\addtolength{\tabcolsep}{-2pt}
\footnotesize
\centering
\begin{tabular}{lrrrrrr}
\toprule
& \multicolumn{4}{c}{\ftl\ $\uparrow$}\\ 
\cmidrule(lr){2-5}
& \multicolumn{2}{c}{BCMS} & \multicolumn{2}{c}{DNS} & \multicolumn{2}{c}{\zsl\ $\uparrow$}\\
\cmidrule(lr){2-3}\cmidrule(lr){4-5}\cmidrule(lr){6-7}
Method & Dev & Test & Dev & Test & BCMS & DNS \\
\midrule
SotA / Rand & --- & $^?$\second{.995} &   --- & ---  &  $^\ddag$.311 & $^\ddag$.351 \\
GlotLID & $^\ddag$.692 & $^\ddag$.697 & $^\ddag$.932 & $^\ddag$.931 & --- & --- \\
FastText & .992 & $^\dag$.983 & $^\ddag$.957 & $^\ddag$.949 & --- & --- \\
\vada & .992 & .992 & .962 & $^\dag$.957 & $^\ddag$.604 & $^\dag$.822 \\ \midrule
\gadas & \second{.993} & \second{.995} & \first{.964} & \first{.962} & \first{.640} & $^\dag$\second{.826} \\
\gadaw & \first{.994} & \first{.997} & \first{.964} & \second{.961} & $^\dag$\second{.631} & \first{.875} \\
\bottomrule
\end{tabular}
\caption{Results on out-of-domain fine-tuned language identification (\ftl) and zero-shot language identification (\zsl). Measure for \ftl\ and BCMS: macro-average F1-score (for comparability); measure elsewhere: prediction accuracy. For \ftl\ and BCMS, the first row shows the current state-of-the-art performance \cite{Rupnik.2023}. For \zsl, the first row shows random performance. See Table~\ref{tab:ftg} for an explanation of the symbols used in the table.}  
\label{tab:ftl-ood}
\squeezeup
\end{table}

\vspace{1.2mm}
\noindent \textbf{Fine-tuned language identification.} The geoadapted PLMs are best at identifying the language in which a text is written: both \gadas\ and \gadaw\ consistently show a higher accuracy than \vada\ (e.g., \gadaw\ vs.\ \vada: +5\% on BCMS dev, +1.9\% on EUR test), and the difference in performance is statistically significant in six out of eight cases (see Table~\ref{tab:ftl}). As opposed to the two geolocation tasks where uncertainty weighting (\gadaw) clearly leads to better results than summing the losses (\gadas), the difference is less pronounced for \ftl\ and significant only in one case (EUR test), even though \gadaw\ numerically outperforms \gadas\ overall. Compared to the language identification models operating on the level of character $n$-grams (GlotLID, FastText), geoadaptation always brings statistically significant performance gains. Even \vada\ outperforms GlotLID and FastText in all cases, indicating that PLMs are generally competitive with more traditional systems on this task. We further notice that the relative disadvantage is particularly pronounced for GlotLID on BCMS. Upon inspection,
we find that GlotLID's inferior performance on BCMS is due to the fact that it predicts more than 80\% of the examples as Croatian. This imbalance can be explained as a result of the domain difference between GlotLID's training data and the \ftl\ evaluation data: while GlotLID was mostly trained on formal texts such as Wikipedia articles and government documents \citep{Kargaran.2023}, we test it on data from Twitter. Crucially,
while Croatian is the only BCMS language that consistently uses Latin script in formal contexts, with Cyrillic script being preferred especially in Serbian,
Latin script is everywhere much more common on social media, even in Serbia \citep{George.2019}. GlotLID seems to be 
heavily affected by this script mismatch and is only very rarely able to correctly predict the language of non-Croatian posts written in Latin script.

These trends are also reflected by the results on the out-of-domain language identification benchmarks: geoadaptation always outperforms adaptation based on language modeling alone as well as models operating on the level of character $n$-gram (see Table~\ref{tab:ftl-ood}). On the SETimes benchmark (BCMS), \gadaw\ further establishes a new state of the art, almost halving the error rate from 0.5\% to 0.3\%. Similarly to in-domain \ftl, the two geoadaptation variants perform similarly. GlotLID again predicts many non-Croatian examples in Latin script as Croatian, leading to a substantially worse performance on BCMS.

The superior performance of the geoadapted models in language identification -- a task that is distinct from geolocation prediction and not typically addressed by means of PLMs --
suggests that the geolinguistic knowledge acquired during geoadaptation is highly generalizable, making it beneficial for a broader set of tasks with a connection to geography, and not only the task used as an auxiliary objective for geoadaptation itself. 

\vspace{1.2mm}
\noindent \textbf{Zero-shot language identification.} Here, the PLMs have to predict the \textit{token} corresponding to the language in which a text is written, e.g., \textit{hrvatski} (`Croatian'). This task requires generalization on two levels: first (similarly to \ftl), the PLMs have not been trained on language identification and are thus required to draw upon the geolinguistic knowledge they have formed during geoadaptation; second (similarly to \zsg), the geolinguistic knowledge has not been provided to them in a form that would make it readily usable in a zero-shot setting -- recall that the geographic information is presented in the form of longitude-latitude pairs (i.e., two scalars), whereas the language modeling head (which is used for the zero-shot predictions) is not trained differently than for vanilla adaptation (\vada). Despite these challenges, we find that geoadaptation substantially improves the performance of the PLMs on \zsl\ (see Tables~\ref{tab:ftl} and \ref{tab:ftl-ood}). The fact that the performance gains are equally pronounced on in-domain (e.g., \gadaw\ vs.\ \vada: +4.2\% on DNS) and out-of-domain examples (e.g., \gadaw\ vs.\ \vada: +5.3\% on DNS) highlights again that geoadaptation endows PLMs with knowledge that allows for a high degree of generalization.

\begin{figure*}[t!]
        \centering      
        \begin{subfigure}[b]{0.22\textwidth}  
          
            \includegraphics[width=\textwidth]{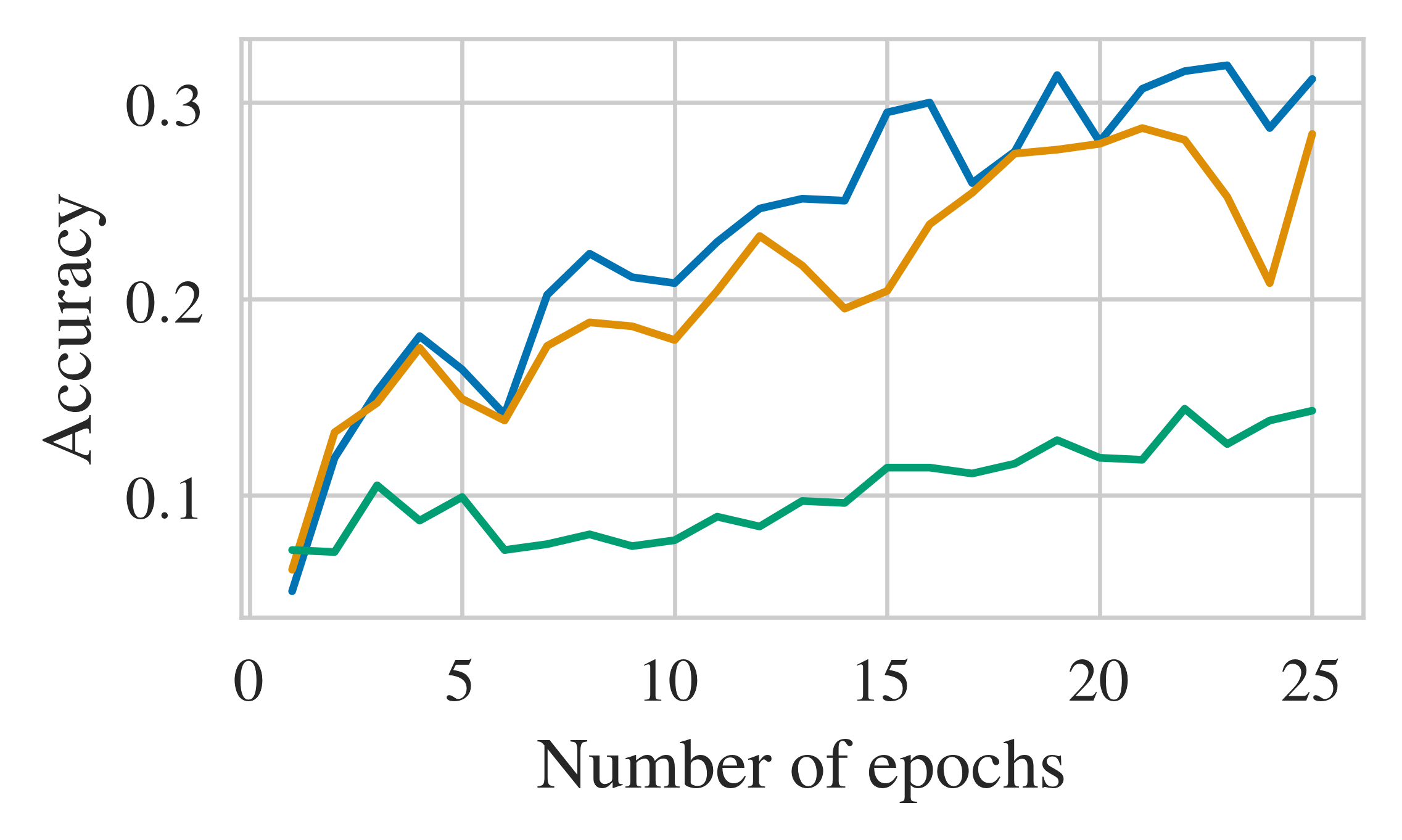}
            \caption[]%
            {{\small \zsg}}    
        \end{subfigure}     
        \begin{subfigure}[b]{0.22\textwidth}   
         
            \includegraphics[width=\textwidth]{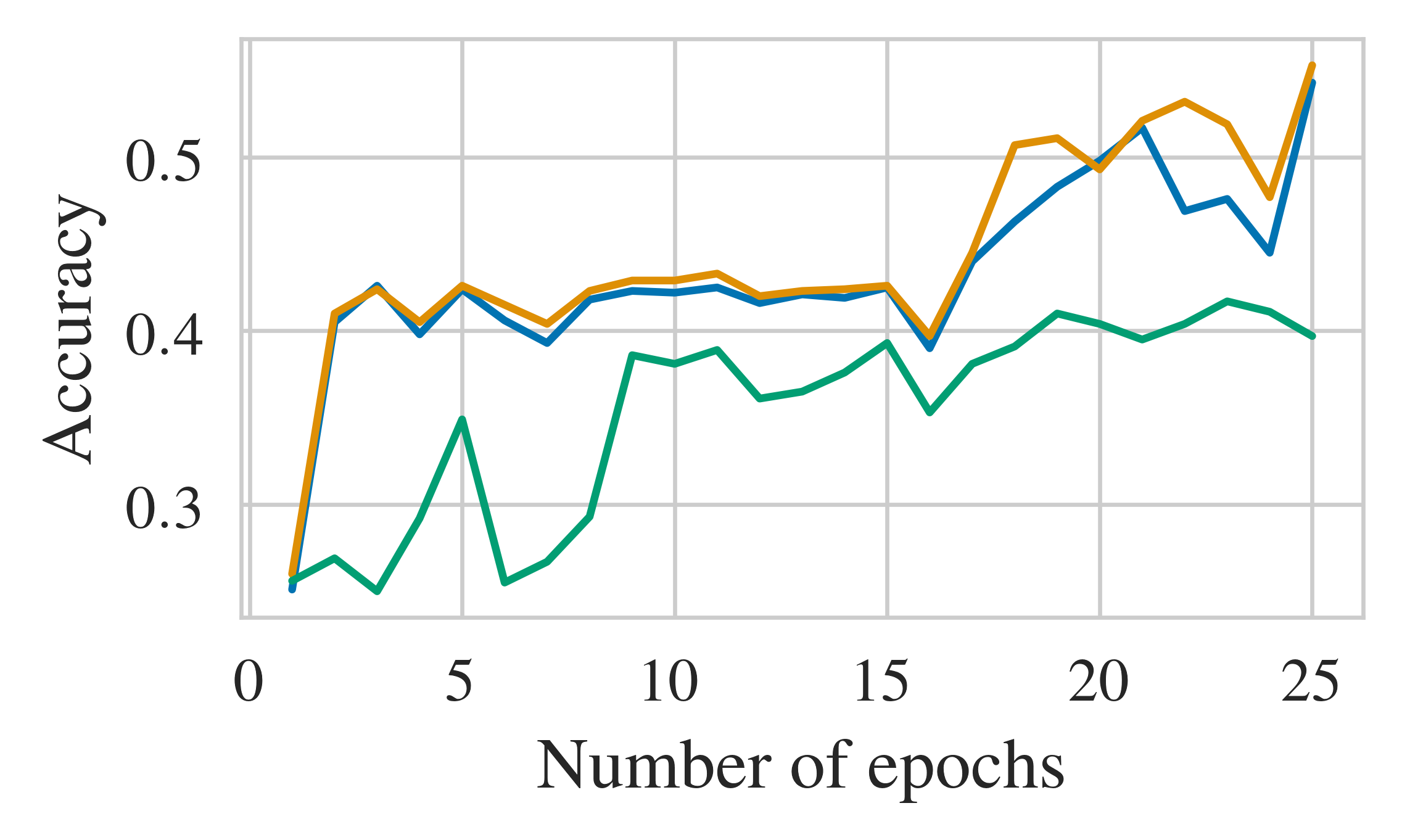}
            \caption[]%
            {{\small \zsl}  }  
        \end{subfigure}
               \begin{subfigure}[b]{0.22\textwidth}   
         
            \includegraphics[width=\textwidth]{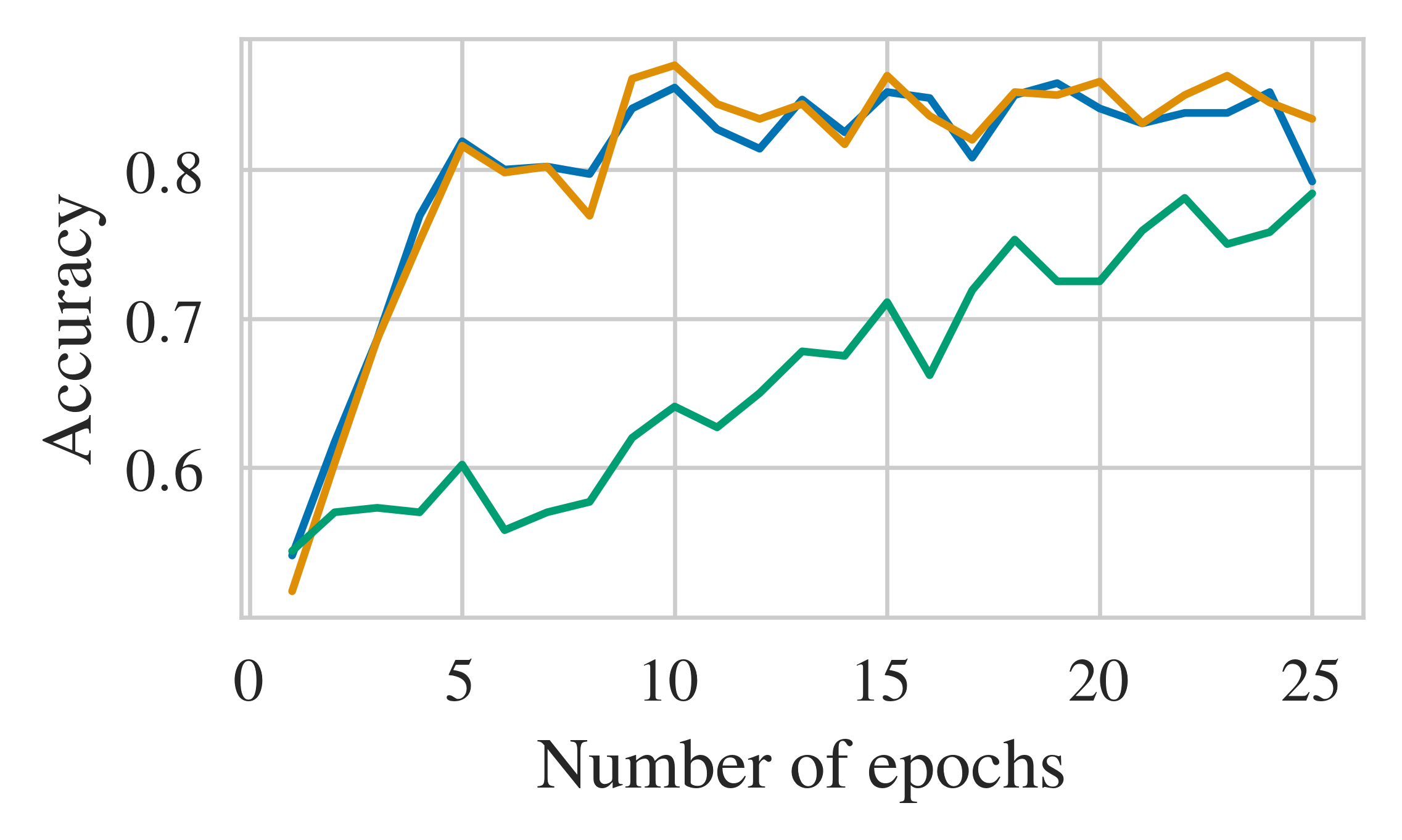}
            \caption[]%
            {{\small \zsd\ Phon}  }  
        \end{subfigure}
               \begin{subfigure}[b]{0.22\textwidth}   
         
            \includegraphics[width=\textwidth]{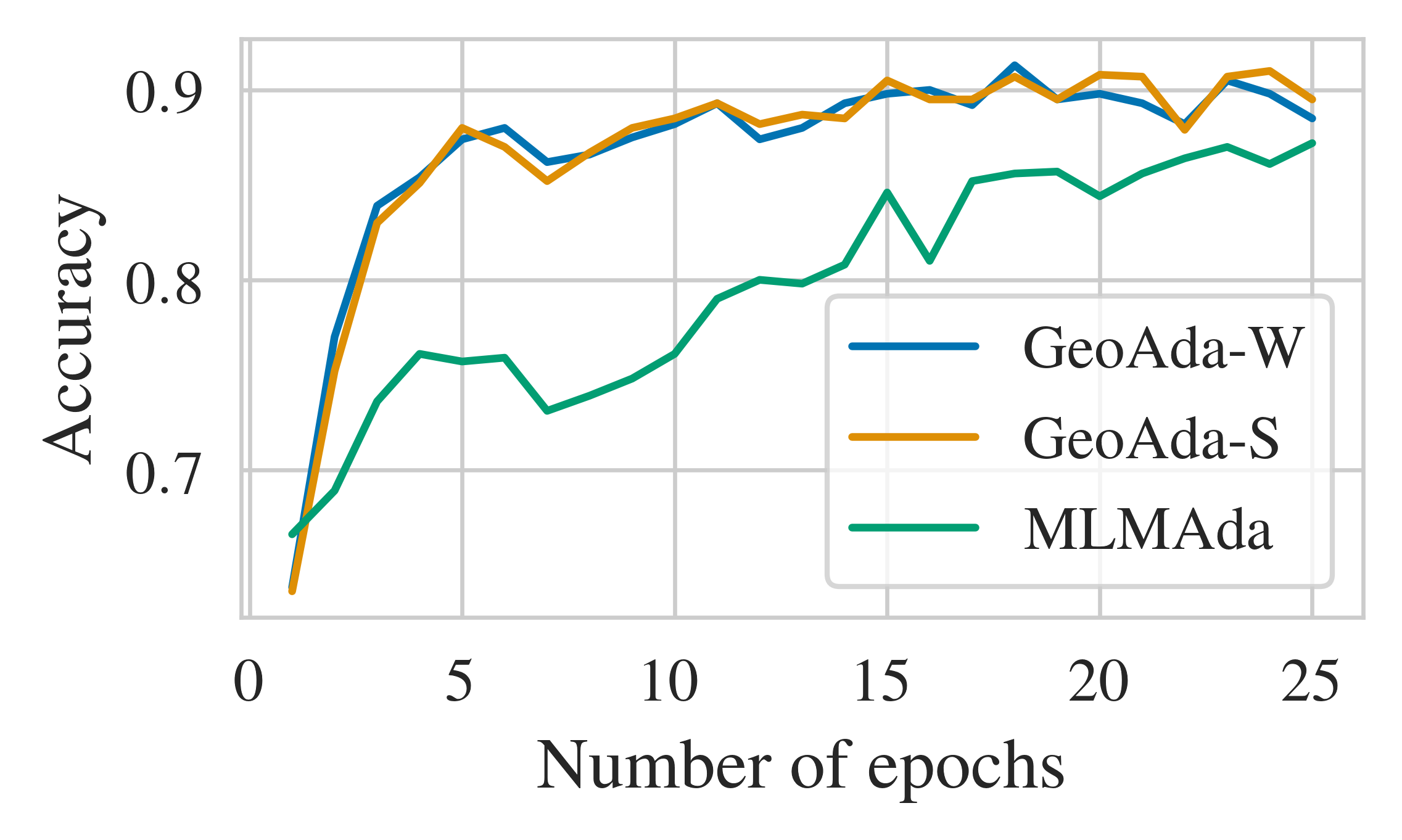}
            \caption[]%
            {{\small \zsd\ Lex}  }  
        \end{subfigure}
        \caption[]{Performance on BCMS \zsg\ (a), \zsl\ (b), and \zsd\ (c, d) for different number of epochs. In stark contrast to geoadaptation (\gadas, \gadaw), language modeling adaptation alone (\vada) barely helps in acquiring geographic knowledge (a), which is also reflected by the consistently worse performance on \zsl\ (b). \vada\ does form dialectal associations after several epochs, but the inductive bias of geoadaptation
allows \gadas\ and \gadaw\ to establish those associations more quickly (c, d).}
        \label{fig:zeroshot}
        \squeezeup
\end{figure*}

\vspace{1.4mm}
\noindent \textbf{Zero-shot dialect feature prediction.} The results on \zsd\ --
phonological (Phon) and lexical (Lex) --
generally follow the trends from the other four tasks (see Table~\ref{tab:zsd}): the
geoadapted PLMs clearly (and statistically significantly) outperform \vada, albeit with
overall narrower margins than in most other zero-shot tasks for BCMS (e.g., \gadas\ vs.\ \vada: +8.6\% on Phon, \gadaw\ vs.\ \vada: +4.1\% on Lex). \vada\ is expectedly more
competitive here: selecting the
word variant that better fits into the linguistic context is
essentially a language modeling task, for which additional language modeling training intuitively helps. For example, typical future
tense constructions in Serbian vs.\ Croatian
(\textit{ja \underline{\'{c}u da okupim}}
vs.\ \textit{ja \underline{\'{c}u okupiti}}, `I'll gather')
have strong selectional preferences on subsequent lexical
units (\citealp{Alexander.2006}; e.g., \textit{porodicu} vs.\
\textit{obitelj} for `family').

\begin{table} [t!]
	\footnotesize
\centering
\begin{tabular}{lrr}
\toprule
& \multicolumn{2}{c}{\zsd\ $\uparrow$}\\
\cmidrule(lr){2-3}
Method & Phon  & Lex \\
\midrule
Rand &  $^\ddag$.501 & $^\ddag$.499  \\
\vada & $^\ddag$.784 & $^\ddag$.872 \\ \midrule
\gadas &  \textbf{.870} & \underline{.910} \\
\gadaw &  \underline{.858} & \textbf{.913} \\
\bottomrule
\end{tabular}
\caption{Results on zero-shot dialect feature prediction (\zsd), which is only conducted for BCMS. Measure: prediction accuracy. See Table~\ref{tab:ftg} for an explanation of the symbols used in the table.}  

\label{tab:zsd}
\squeezeup
\end{table}

\begin{figure}[t!]
        \centering      
        \begin{subfigure}[b]{0.23\textwidth}  
          
            \includegraphics[width=\textwidth]{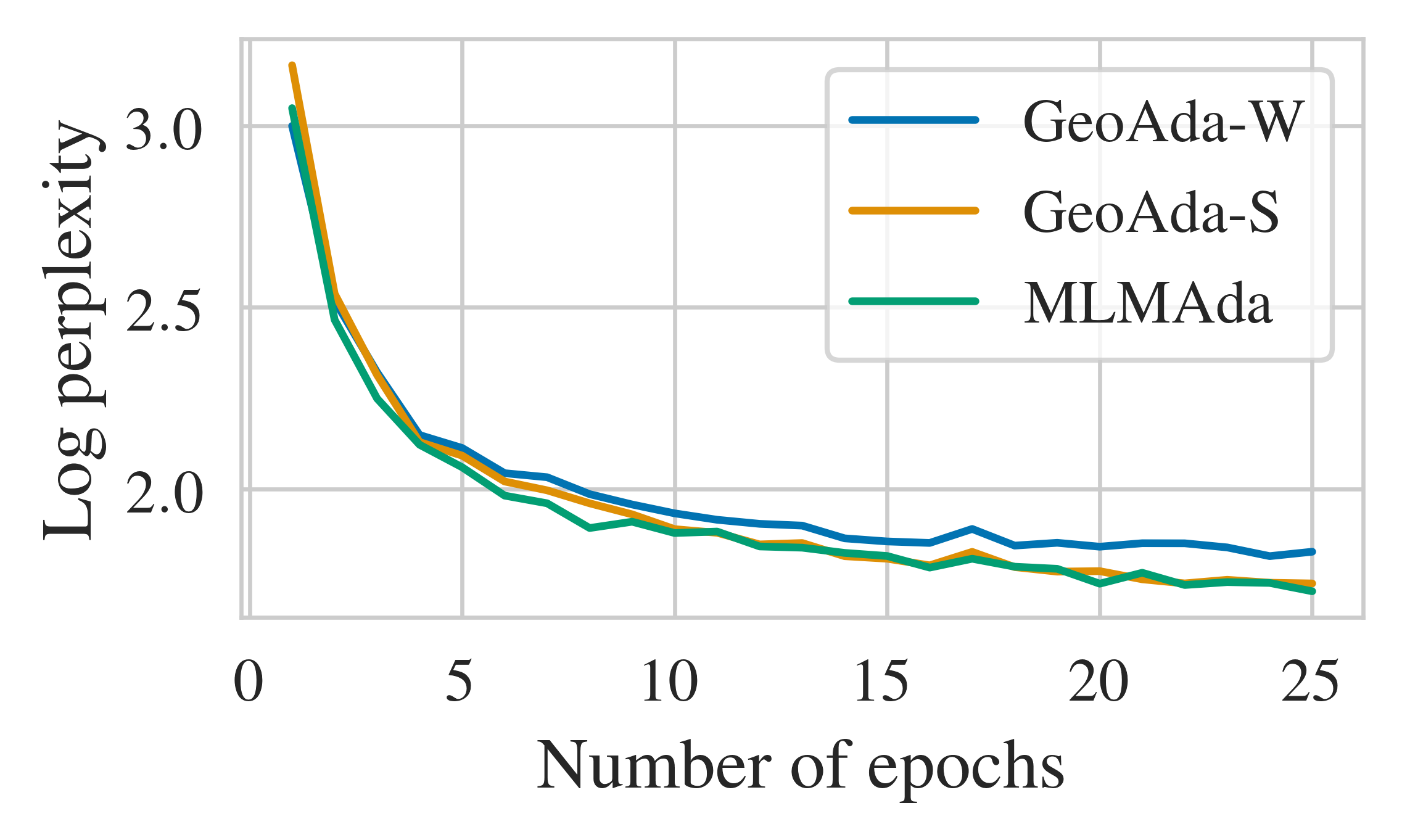}
            \caption[]%
            {{\small Language modeling}}    
        \end{subfigure}     
        \begin{subfigure}[b]{0.23\textwidth}   
         
            \includegraphics[width=\textwidth]{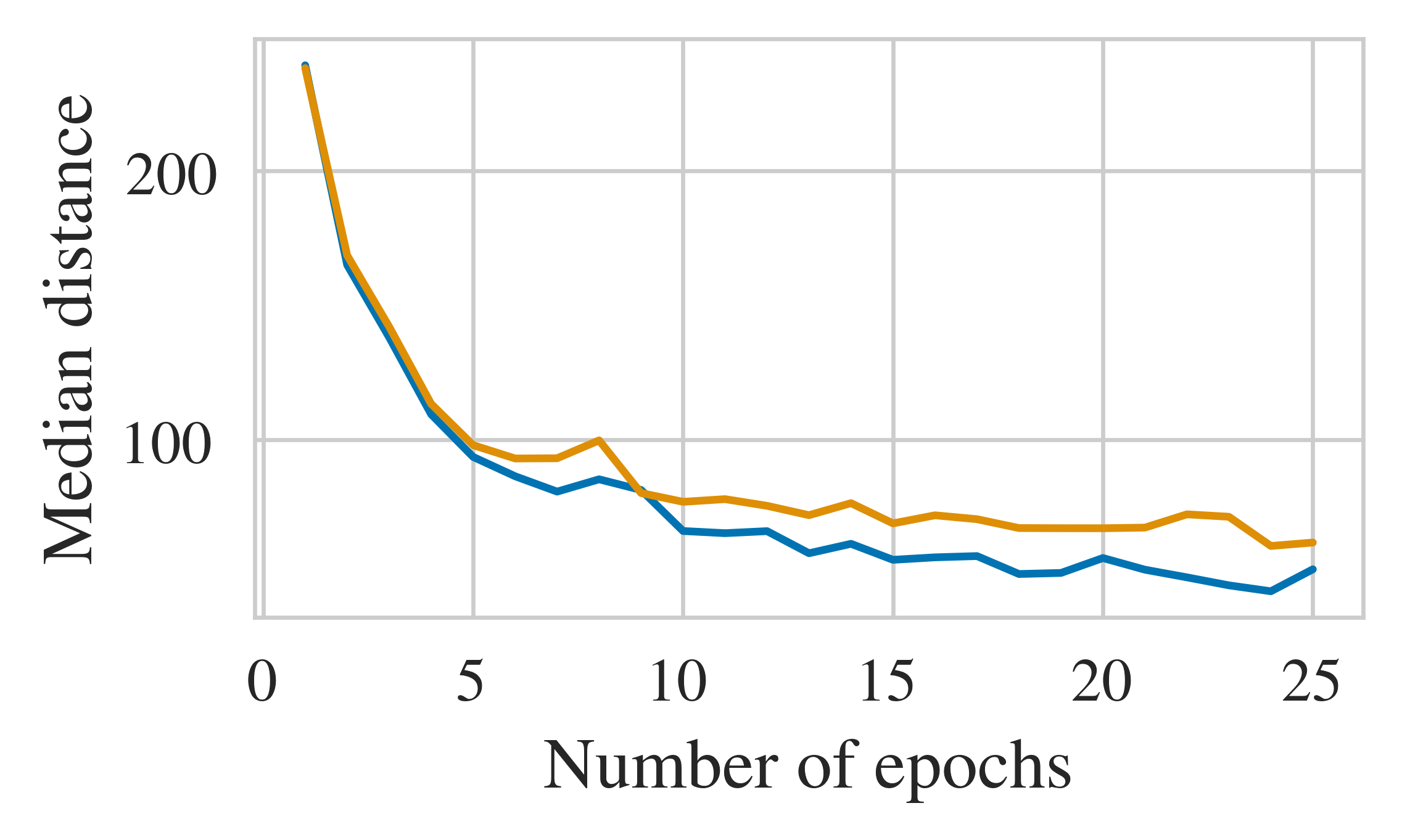}
            \caption[]%
            {{\small Geolocation prediction}}   
        \end{subfigure}
        \caption[]{(Geo-)adaptation diagnostics. The figure illustrates how log perplexity of language modeling (a) and median distance of token-level geolocation prediction (b) change on dev during BCMS geoadaptation. }
        \label{fig:gepadaptation}
        \squeezeup
\end{figure}

We further verify this by comparing the
zero-shot performance on BCMS for different model checkpoints obtained during training. The performance curves over 25 (geo-)adaptation
epochs, shown in Figure \ref{fig:zeroshot}, confirm our
hypothesis that longer language modeling adaptation substantially improves
the performance of \vada\ on predicting
dialect features, but its
benefits for geolocation prediction and language identification remain limited. While prolonged language modeling adaptation allows \vada\ to eventually learn the
dialectal associations, the inductive bias of the knowledge injected via geoadaptation
allows \gadas\ and \gadaw\ to reach high performance much sooner, after merely two to three epochs.

\begin{table*} [t!]
	\footnotesize
\centering
\begin{tabular}{lrrrrrrrr}
\toprule
& \multicolumn{2}{c}{\ftg\ $\downarrow$} & & \multicolumn{2}{c}{\ftl\ $\uparrow$} & & \multicolumn{2}{c}{\zsd\ $\uparrow$}\\ 
\cmidrule(lr){2-3}\cmidrule(lr){5-6}\cmidrule(lr){8-9}
Model & Dev  & Test & \zsg\ $\uparrow$ & Dev & Test & \zsl\ $\uparrow$ & Phon & Lex\\
\midrule
GeoAda-Seq & $^\dag$27.35 & \first{12.13} & $^\dag$.188 & .737 & .730 & $^\dag$.542 & $^\dag$.844 & $^\dag$.885\\
GeoAda-Tok & \first{23.90} & \first{12.13} & \first{.319} & \first{.743} & \first{.734} & \first{.553}& \textbf{.870} & \textbf{.913}\\
\bottomrule
\end{tabular}
\caption{Comparison between sequence-level geoadaptation (GeoAda-Seq) and token-level geoadaptation (GeoAda-Tok) for BCMS. GeoAda-Tok stands for the better-performing model between \gadas\ and \gadaw\ on each task (see Tables \ref{tab:ftg}, \ref{tab:ftl}, and \ref{tab:zsd}). See Table~\ref{tab:ftg} for an explanation of the symbols used in the table.}  \label{tab:results-ablation}
\end{table*}

\vspace{1.4mm}
\noindent\textbf{Effects of loss weighting.} The dynamic weighting of $\mathcal{L}_{\mlm}$ and
$\mathcal{L}_{\geo}$ (i.e., \gadaw) clearly outperforms the simple summation of
the losses (i.e., \gadas) on the geolocation prediction
tasks (\ftg, \zsg), but the difference between the two geoadaptation variants is less pronounced for \ftl, \zsl, and \zsd. While
geographic knowledge is beneficial for all five tasks, geolocation prediction arguably demands a more direct
exploitation of that knowledge. Comparing
the model variants in terms of the two
task losses, we observe
that \gadas\ reaches lower $\mathcal{L}_{\mlm}$ levels,
whereas \gadaw\ ends with lower $\mathcal{L}_{\geo}$ levels (see Figure \ref{fig:gepadaptation} for the example of BCMS), which would explain the differences in their performance. We inspect \gadaw's task uncertainty weights after geoadaptation and observe $\eta_{\mlm} = 0.29$ and  $\eta_{\geo} = -0.35$ for AGS, $\eta_{\mlm} = 1.12$ and  $\eta_{\geo} = -1.22$ for BCMS, $\eta_{\mlm} = 0.84$ and $\eta_{\geo} = -1.23$ for DNS, and $\eta_{\mlm} = 0.90$ and $\eta_{\geo} = -1.95$ for EUR. Thus, \gadaw\ consistently assigns more importance to $\mathcal{L}_{\geo}$.\footnote{Because $\tilde{\mathcal{L}}_l \propto - \eta_l$
(see Equation \ref{eq:eta_weight}), the smaller the value of $\eta_l$, the larger the emphasis on task $l$.} The fact that the divergence of the task uncertainty weights is smallest for AGS explains why the difference between \gadas\ and \gadaw\ on \ftg/\zsg\ is least pronounced for that language group.

\begin{figure}[t!]
        \centering
        \includegraphics[width=0.75\linewidth]{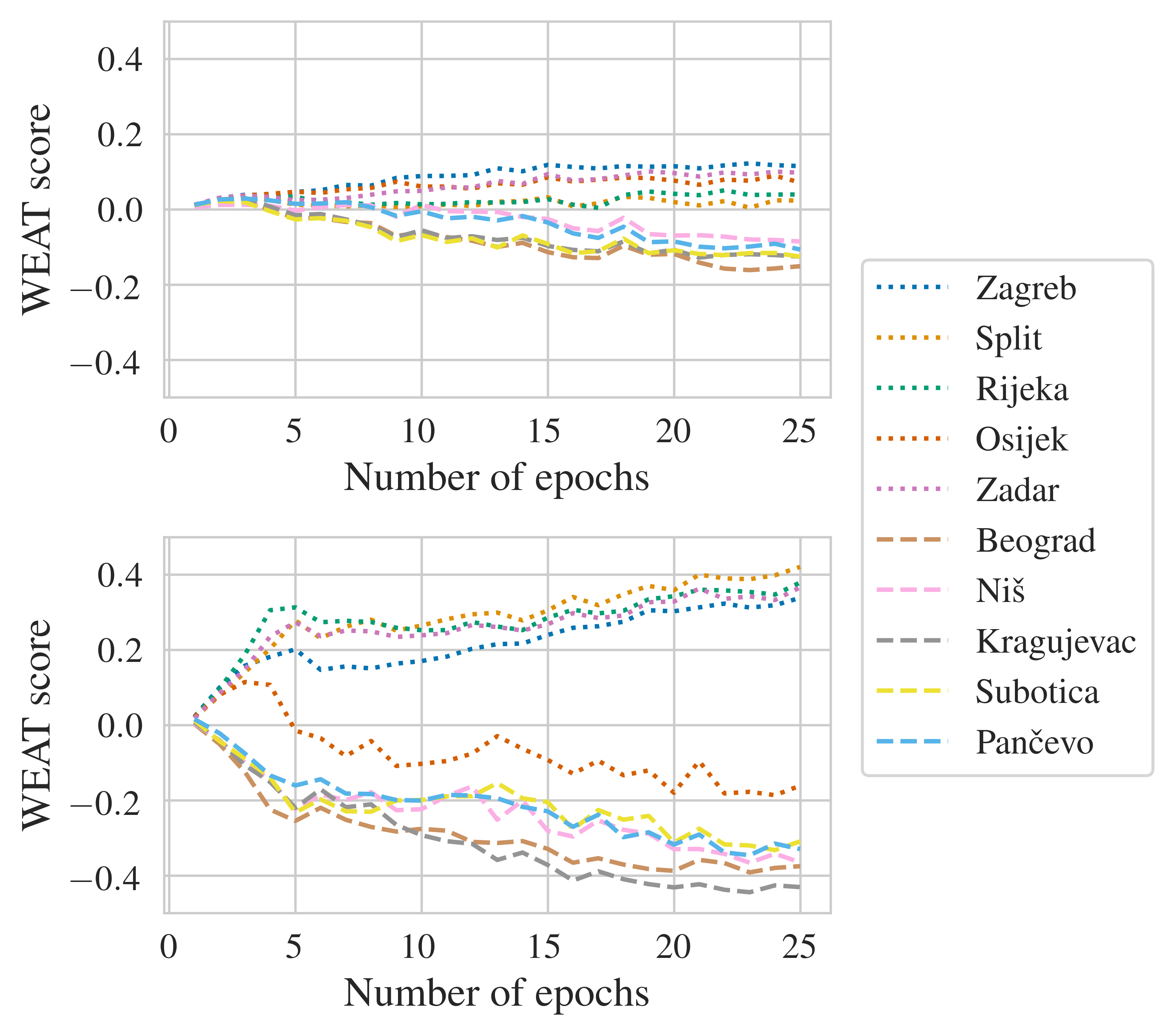}
        \caption{Association strength between the BERTi\'c embeddings of Croatian/Serbian cities and \textit{ije}/\textit{e} variants for \vada\ (top) and \gadaw\ (bottom), measured using WEAT \citep{Caliskan.2017}. A positive or negative score indicates that a city is associated more strongly with the \textit{ije} or \textit{e} variants, respectively.}
        \label{fig:WEAT_score}
\squeezeup
\end{figure}

\vspace{1.4mm}
\noindent \textbf{Sequence-level geoadaptation.} The decision to inject geographical information at the level of tokens was motivated by the central importance of the lexicon for geographically-conditioned linguistic variability (see \S\ref{sec:related-work}).
A plausible alternative -- one less tied to lexical variation alone -- is to geoadapt the PLMs by predicting the geolocation from the representation of the whole input text, i.e., to feed the contextualized representation of the \texttt{[CLS]} token to the regressor that predicts longitude and latitude. For comparison, we evaluate this variant too (GeoAda-Seq) and compare it against the best token-level geoadapted model (GeoAda-Tok; e.g., GeoAda-W for BCMS \ftg) on all PLMs and tasks. For reasons of space, we only present BCMS here, but the overall trends for AGS, DNS, and EUR are very similar.

Sequence-level geoadaptation trails token-level geoadaptation on all tasks except for fine-tuned geolocation prediction (see Table \ref{tab:results-ablation}). In general, while the difference is small for the fine-tuned tasks, it is large (and always significant) for the zero-shot tasks -- for example, GeoAda-Seq performs only slightly
better than \vada\ on \zsg\ (see Table~\ref{tab:ftg}). This suggests that injecting geographic information on the level of tokens allows the PLMs to acquire more nuanced geolinguistic knowledge. Nonetheless, sequence-level geoadaptation still outperforms the non-geoadapted baselines.

\vspace{1.4mm}
\noindent \textbf{Geoadaptation as geographic retrofitting.} Even though it makes intuitive sense that minimizing $\mathcal{L}_{\geo}$ improves the geolinguistic knowledge of PLMs, we want to determine the exact mechanism by which it does so. Based on the results described so far, we make the following hypothesis: geoadaptation changes the representation space of the PLMs in such a way that tokens indicative of a certain location are brought close to each other, i.e., it has the effect of \emph{geographic retrofitting} \citep{Hovy.2018}. We examine this hypothesis by analyzing (i) how the representations of toponyms and lexical variants change in relation to each other, and (ii) how the representations of toponyms change internally. We examine the PLM output embeddings (which directly impact the zero-shot predictions) and focus on BCMS.

\begin{figure}[t!]
        \centering
        \includegraphics[width=0.85\linewidth]{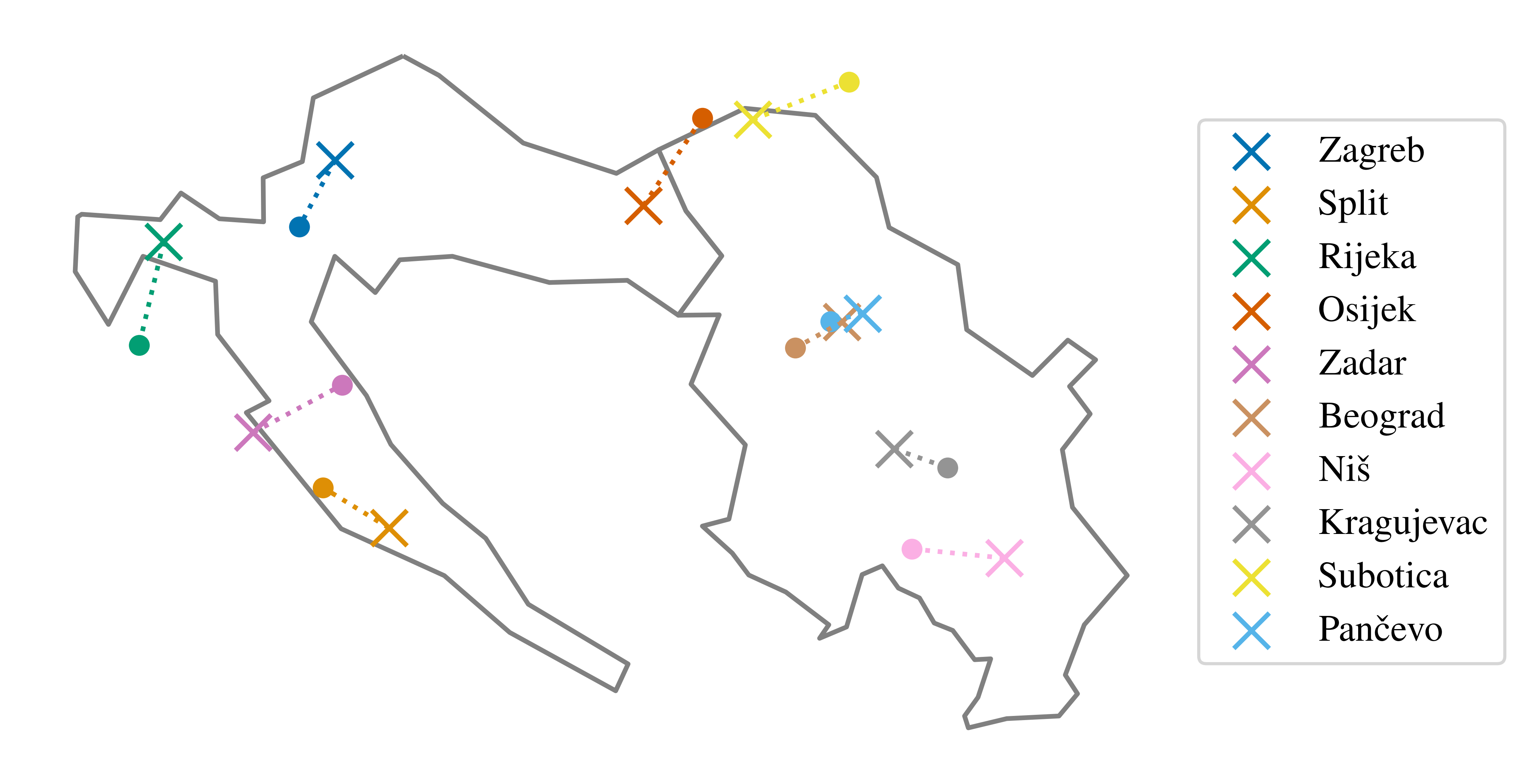}  
        \caption{The first two principal components of the city output embeddings (points), plotted on top of a geographic map of Croatia and Serbia. The x-marks indicate the actual geographic locations of the cities.}
        \label{fig:pca_vs_map}
\squeezeup
\end{figure}

For the first question, we use the geoadaptation data to compute type-level embeddings for the five largest Croatian (\textit{Zagreb}, \textit{Split}, \textit{Rijeka}, \textit{Osijek}, \textit{Zadar}) and Serbian (\textit{Beograd}, \textit{Niš}, \textit{Kragujevac}, \textit{Subotica}, \textit{Pančevo}) cities as well as the \textit{ije}/\textit{e} variants used for \zsd. Following established practice \citep[e.g.,][]{vulic2020probing,litschko2022cross}, we obtain type-level vectors for words (i.e., city name or phonological variant) by averaging the contextualized output representations of their token occurrences. 
We then resort to WEAT \citep{Caliskan.2017}, a measure that quantifies the difference in association strength between a word (in our case, a city name) and two word sets (in our case, \textit{ije} vs.\ \textit{e} phonological variants). A positive or negative score indicates that a city name is associated more strongly with the \textit{ije} or \textit{e} variants, respectively. 
Figure~\ref{fig:WEAT_score} shows that during geoadaptation (\gadaw), the Croatian city names develop a strong association with the \textit{ije} variants (i.e., positive WEAT scores), whereas the Serbian city names develop a strong association with the \textit{e} variants (i.e., negative WEAT scores), which is exactly in line with their geographic distribution \citep{Alexander.2006}. By contrast, the associations created during adaptation based on language modeling alone (\vada) are substantially weaker.

We then use the same set of 10 Croatian and Serbian cities and compare their pairwise geodesic distances against the pairwise cosine distances of the city name embeddings, at the end of geoadaptation. The correlation between the two sets of distances (Pearson's $r$) is only 0.577 for \vada, but 0.881 for \gadaw, indicating almost perfect correlation. Furthermore, after only five epochs, the correlation is already 0.845 for \gadaw\ (vs.\ only 0.124 for \vada).
This striking correspondence between real-world geography and the topology of the embedding space of geoadapted PLMs can also be seen by plotting the first two principal components of the city name embeddings on top of a geographic map, where we use orthogonal Procrustes \citep{Schonemann.1966, Hamilton.2016} to align the points (see Figure~\ref{fig:pca_vs_map}).

These results are strong evidence that geoadaptation indeed acts as a form of geographic retrofitting. The geographically restructured representation space of the PLMs can then be further refined via fine-tuning (as in \ftg\ and \ftl) or directly probed in a zero-shot manner (as in \zsg, \zsl\, and \zsd).
\section{Conclusion}

We introduce geoadaptation, an approach for task-agnostic
continued pretraining of PLMs that forces them to learn
associations between linguistic phenomena and geographic
locations. The method we propose for geoadaptation
couples language modeling and token-level geolocation
prediction via multi-task learning. While we focus on PLMs pretrained via masked language modeling, geoadaptation can in principle be applied to autoregressive PLMs as well.
We geoadapt four PLMs and obtain 
consistent gains on five tasks, establishing a new state of the art on established benchmarks.
We further show that geoadaptation acts as a form of geographic retrofitting. Overall, we see our study as an exciting step
towards NLP technology that takes into
account extralinguistic aspects in general and
geographic aspects in particular.

\section*{Acknowledgements}

This work was funded by the European Research Council (grant \#740516 awarded to LMU Munich) and the Engineering and Physical Sciences Research Council (grant EP/T023333/1 awarded to University of Oxford). Valentin Hofmann was also supported by the German Academic Scholarship Foundation. Goran Glava\v{s} was supported by the EUINACTION grant from NORFACE Governance and German Science Foundation (462-19-010, GL950/2-1). Nikola Ljube\v{s}i\'c was supported by the Slovenian Research and Innovation Agency (P6-0411). We thank the reviewers and action editor for their very helpful comments.

\bibliography{0-main}

\bibliographystyle{acl_natbib}

\end{document}